\documentclass[final,5p,times,twocolumn]{elsarticle}

\usepackage{amssymb}
\usepackage{xcolor}
\usepackage{caption}
\usepackage{subcaption}
\usepackage{hyperref}
\usepackage{graphicx,subcaption}
\usepackage{amsmath}
\usepackage{url}
\journal{Nuclear Physics B}

\begin{document}

\begin{frontmatter}



\title{Landcover classification and change detection using remote sensing and machine learning: a case study of Western Fiji}


\author[inst3]{Yadvendra Gurjar}

\author[inst1]{Ruoni Wen}

\author[inst2]{Ehsan Farahbakhsh}
\author[inst1]{Rohitash Chandra}
\affiliation[inst1]{Transitional Artificial Intelligence Research Group, School of Mathematics and Statistics, UNSW Sydney, Sydney, Australia}

\affiliation[inst2]{School of Geosciences, University of Sydney,, Sydney, Australia}

% \affiliation[inst3]{Centre for AI and Innovation, Pingla Institute,, Sydney, Australia}

\affiliation[inst3]{Department of Electrical Engineering,, Indian Institute of Technology Delhi, Delhi, India}

\begin{abstract}
As a developing country, Fiji is facing rapid urbanisation, which is visible in the massive development projects that include housing, roads, and civil works.  In this study, we present machine learning and remote sensing frameworks to compare land use and land cover change from 2013 to 2024 in Nadi, Fiji. The ultimate goal of this study is to provide technical support in land cover/land use modelling and change detection. We used Landsat-8 satellite image for the study region and created our training dataset with labels for supervised machine learning. We used  Google Earth Engine and unsupervised machine learning via k-means clustering to generate the land cover map. We used convolutional neural networks to classify the selected regions' land cover types. We present a visualisation of change detection, highlighting urban area changes over time to monitor changes in the map.

%The classification accuracies for all three models were high. ANN had an accuracy of 94.857\%, RF had an accuracy of 96.952\%, and CNN had an accuracy of 99.05\%. 
\end{abstract}




\end{frontmatter}



% \textcolor{red}{main text}

\section{Introduction}
\label{sec:sample1}

Land-use and land cover-change (LULCC) describe the physical state of the land surface and the utilisation of land and its resources by humans \cite{Hasmadi2009}. It is a dynamic process that describes temporal and spatial changes \cite{wang2022machine} with various applications. LULC data has been analysed to assess vegetation health and variations in Land Surface Temperature (LST) in the Bisalpur wetland buffer zone from 2013 to 2024 \cite{singh2024dynamics}. Singh et al. \cite{singh2023classification} classified vegetation in the mountainous regions of the Indian Western Himalayas, an area characterised by its mega-diversity, wide climate variation, diverse vegetation types, and complex topography. Tu et al. \cite{Tu2020} have used land cover maps to detect the distribution of various land covers in Guangdong, one major province in China.

Remote sensing provides the technology to observe areas from space (space-borne) without direct contact \cite{aggarwal2004principles}, which has advantages in providing multi-spectra and multi-temporal data compared to manual surveys \cite{rundquist2001review}. Such technology has been used in acquiring land surface images for LULCC mapping. The \textit{light detecting and ranging} (LiDAR) and \textit{synthetic aperture radar} (SAR) systems are examples of active sensors in satellites \cite{montesano2014uncertainty}.  The \textit{Landsat} satellite is an example of a passive sensor that detects the reflected radiation of natural sunlight \cite {stamatiadis2010comparison}, where each pixel in its image typically contains the spectral measurement of reflectance \cite{wang2022machine}. Due to the working function of both types of sensors, passive sensors can be affected by haze from clouds \cite{lechner2020applications}, and the active sensor can penetrate clouds and smoke \cite{lechner2020applications}. However, in radar images, common types are speckle noise \cite{maity2015comparative} and salt and pepper noise \cite{ali2022design}.

Change detection compares the land surface in the same study area at different times based on the same LULCC \cite{chen2002quantification}. The map-to-map approach compares \textcolor{black}{land cover images}\cite{giri2012remote} and provides details such as the location of the changed areas and how land covers were replacing each other. The map-to-map comparison method is more prevalent in detecting change \cite{pouliot201212}. Besides, standardisation is crucial for dealing with multi-temporal images \cite{aplin2004remote}. The action includes performing atmospheric correction \cite{lu2002assessment}, obtaining images from a single source \cite{kastens2002time}, and data conversion if using multiple image sources \cite{aplin2004remote}.

Machine learning  models in combination with remote sensing has been prominent in LULCC using the spatial-spectral features of satellite data \cite{wang2022machine}. \textcolor{black}{The random forest is an example of a machine learning model that has been widely used in land cover classification due to its accuracy\cite{shi2016assessment,avci2023comparison,le2022rapid,adam2014land,amini2022urban}.} Vasuki et al. \cite{vasuki2019spatial} compared the performances of machine learning models such as support vector machine, random forests and Naïve Bayes for land cover change detection in Del Park mine, Australia. However, given the prominence of deep learning, the remote sensing community has gradually shifted its focus towards deep learning \cite{ma2019deep}.

\textcolor{black}{Convolutional neural network (CNN) are deep learning models widely used in image processing \cite{naranjo2020review,han2020underwater,ehtisham2024computing,kim2018convolutional} and utilised in remote sensing \cite{wang2022machine}.  CNNs utilise spatial context information, leading to outstanding performance in land cover classification \cite{yoo2019comparison}. Pu et al.\cite{pu2019water} used CNN to map the relationship between Landsat8 images and water-quality levels, incorporating a transfer-learning strategy to address data scarcity. Yoo et al. \cite{yoo2019comparison} provided a comparison of classification performance between CNN and random forests for local climate zones in urban areas using Landsat images, and Ma et al. \cite{ma2019deep} explained that CNNs, originally designed to handle data in multiple arrays\cite{lecun2015deep}, are particularly effective for processing multiband remote-sensing images due to their structured pixel arrangement.}

 Fiji is a developing country facing rapid urbanisation  in the last decades with development projects that include housing, road and civil works There was slow growth during COVID-19 lockdowns, which has rebounded according to the World Bank \cite{wbfj}.  Nadi is a major town in western Viti Levu (the main island of Fiji), home to the international airport, and serves as the centre of Fiji's tourism industry. Tourism has been the backbone of the Fijian economy \cite{harrison2013contribution,movono2018fijian}, which overtook the sugar industry \cite{mahadevan2009viability,narayan2005economic} given land disputes \cite{lal2001alta,kumari2016does}.  The tourism and sugar industry faces challenges of climate extremes such as cyclones and floods\cite{jayaraman2018natural} along with management of coastal resources\cite{singh2021coastal}. In terms of development, there are further challenges along with restriction of development of land since 92\% of land is protected as native land \cite{lal2001alta}. A study of Nadi would be particularly interesting due to the developments by the hotel industry \cite{tuisuva2023exploring} and major natural disasters such as flooding \cite{paquette2012flood,chandra2016deconstructing}. This area also features the sugar industry  that faced challenges in previous decades in Fiji due to land tenure problems \cite{moynagh1978land}. Over the last decades, the shift from sugar to the tourism industry has affected the development of Nadi. A large amount of agricultural land which was used for the sugar industry has been gradually developed as a residential zone. An assessment of these using remote sensing via LULCC can give further insights.

 %We used the free open-access Landsat-8 OLI 30m resolution images and created our own data labels for supervised learning. The data pre-processing and labelling were done on the Google Earth Engine cloud platform. We evaluated supervised learning algorithms based on the 2013's satellite image, including Random Forests, Artificial neural networks, and convolutional neural networks. Using the Google Earth Engine build-in function, an unsupervised learning algorithm K-means was also tried. The algorithm that gave the best classification result was used in mapping the LULC in Nadi from 2013 to 2019. We present a visualisation of change detection, highlighting urban area changes over time to monitor changes in the map.

 In this study, we present a  machine learning and remote sensing framework to compare land use and land cover change over a decade (2013-2024) in Nadi, Fiji. The ultimate goal of this study is to provide technical support in land cover/land use modelling and change detection. We used   Landsat-8 satellite image for the study region and created  our own hand-labelled training dataset  for supervised machine learning. We used  Google Earth Engine and an unsupervised machine learning via k-means clustering for generating the land cover map. We used convolutional neural networks for classifying the land cover types of the selected regions. We further compare the results with simple neural networks and random forest models. 

\section{Methods}
\label{sec:sample1}

\subsection{Study area}

\begin{figure*}
  \centering
  \begin{subfigure}[t]{.58\linewidth}
    \centering\includegraphics[width=1\linewidth]{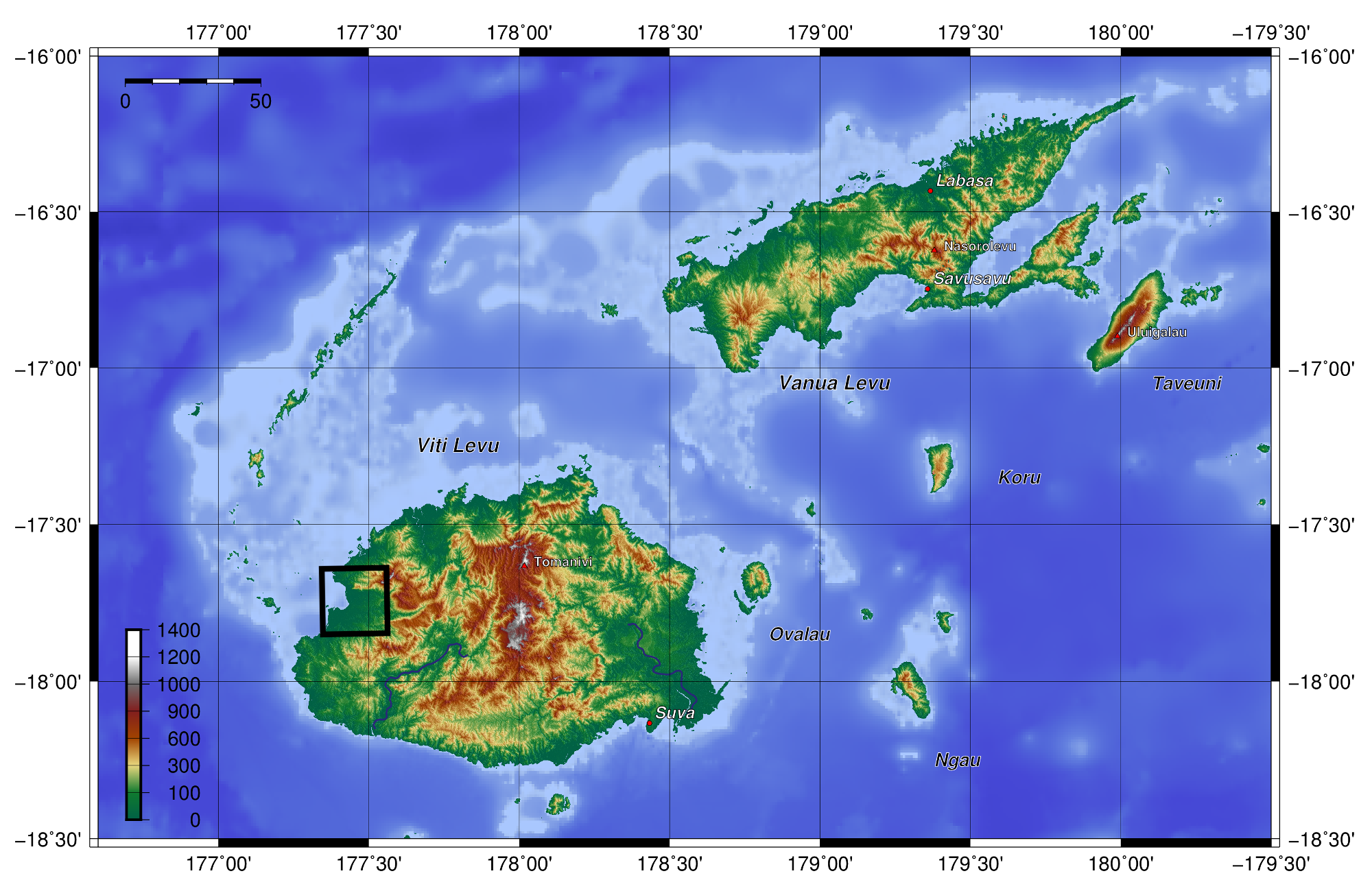}
    \caption{}
  \end{subfigure}\hfill
  \begin{subfigure}[t]{.38\linewidth}
    \centering\includegraphics[width=1\linewidth]{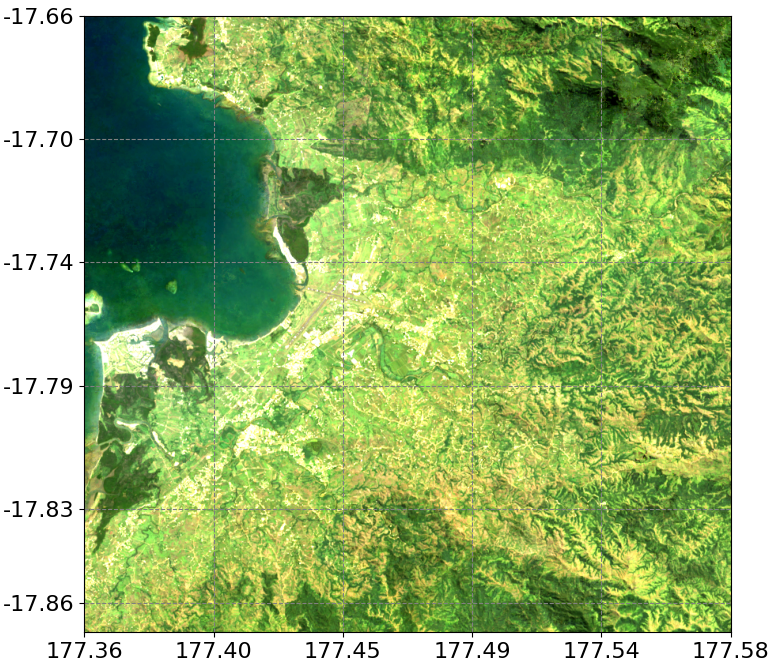}
    \caption{}
      \end{subfigure}\hfill
\caption{\textcolor{black}{Location of the study area in Viti Levu, Fiji\cite{wiki:xxx}. (a) The map of Viti Levu with the study area highlighted in black. (b) A closer view of the selected study area showing its land cover characteristics}}
\label{fiji_topo}
\end{figure*}

The Republic of Fiji is composed of a group of islands in the South Pacific Ocean \cite{Frost1979}. The total area of Fiji covers up to 18,376 km\(^2\), which consists of about 330 islands \cite{mataki2006baseline} of which most are uninhabited and the major population resided on two major islands, known and Viti Levu and Vanua Levu. Viti Levu is the biggest island of Fiji, which harbours over 70 \%  of total population\cite{Gravelle2008}.  The Eastern side experiences heavy rainfall, where the Western side is drier \cite{kumar2014rainfall} and hence home to the sugar industry  and Nadi. This study covers an area of 625km\(^2\)(25 km x 25 km) encompassing the Nadi region (Figure \ref{fiji_topo}). Nadi is known as the third biggest city in Fiji due to its population. The government  has not yet declared Nadi as a city, due to population size and level of development since the region features villages intertwined with sugar and agricultural fields. The Denarau island is the tourism hub of Nadi which features a group of international hotels and resorts which has also  attracted residential development in the last two decades \cite{bernard2015luxury}.
%Among all visitors who travel to Fiji, 21\% stay in Nadi and 21\% on Denarau island around Nadi.

% \begin{figure}[h]
% \includegraphics[width=0.5\textwidth]{Fiji_topo.png}
% \caption{\textcolor{black}{The position of the study area in Viti Levu, Fiji \cite{wiki:xxx}.}  The study area is marked in black.}
% \label{fiji_topo}
% \end{figure}

% \subsection{Data preparation}

% The Landsat 8 OLI (Operational Land Imager ) data has a spatial resolution of 30 meters (m) and 12 bits of radiometric resolution\cite{pahlevan2014orbit,chander2009summary}. Among eleven provided bands, seven bands are used including Ultra (0.43-0.45), Blue (0.45-0.51), Green (0.53-0.59), Red (0.64-0.67), Near Infrared (0.85-0.88), Shortwave Infrared 1 (1.57-1.65), and Short-wave Infrared 2 (2.11-2.29)\cite{Phiri2017}. The LULCC mapping in this study is based on seven median composite images generated from 2013 to 2019. Each image has a size of 780*818, which covers 638040 pixels.
% A projection code is required when the extracted satellite images are transferred from the Google Earth Engine to other platforms, ensuring the consistency of geographic data. In this study, we used a coordinate reference system (EPSG 4326) \footnote{\url{https://epsg.io/4326}} when exporting the images.
% \textcolor{black}{re-write version: A projection code is necessary to maintain geographic data consistency when satellite images are transferred across platforms. This study employed the EPSG 4326 coordinate reference system to ensure accurate alignment during image export.}

\subsection{Data Preparation}

\textcolor{black}{The Landsat 8 Operational Land Imager (OLI) provides multispectral data with a spatial resolution of 30 meters and a radiometric resolution of 12 bits \cite{pahlevan2014orbit, chander2009summary}. Out of the eleven available bands, seven were utilized in this study for land use and land cover change (LULCC) analysis. These include the Ultra (0.43–0.45 µm), Blue (0.45–0.51 µm), Green (0.53–0.59 µm), Red (0.64–0.67 µm), Near-Infrared (0.85–0.88 µm), Short-Wave Infrared 1 (1.57–1.65 µm), and Short-Wave Infrared 2 (2.11–2.29 µm) bands \cite{Phiri2017}.}

\textcolor{black}{For this study, ten median composite images were generated using Google Earth Engine (GEE), each corresponding to a different year between 2013 and 2022. These composites were created to minimize the impact of transient atmospheric disturbances such as clouds. Each image covers an area of 780×818 pixels, totaling 638,040 pixels per image.}

\textcolor{black}{When exporting satellite imagery from GEE to external platforms for further processing, a consistent coordinate reference system is essential to preserve geospatial accuracy. In this work, we used the 'EPSG:4326' geographic coordinate system \footnote{\url{https://epsg.io/4326}} to ensure proper spatial alignment and compatibility with other geospatial datasets.}

% \subsubsection{Cloud removal}

%  In the Google Earth Engine, we selected the 'QA\_PIXEL' band  to access each pixel's bit-packed combination.
%  %not clear what you trying to say, better first tell what is Google Earth Engine and what it gives 

% \textcolor{black}{Google Earth Engine (GEE) is a cloud-based, free-to-use platform \cite{tamiminia2020google} that provides access to a global archive of satellite imagery. It includes the necessary computing power and algorithms for processing such data effectively \cite{kumar2018google}.The adoption of GEE has rapidly expanded across various applications, especially within the remote sensing community \cite{zhao2021progress}.}

% %what is pit-packed combination?

% %what is a bit? Do you mean pixel of data is a bit?
 
%  Each bit (pixel) in the satellite image data extracted from the Google Earth Engine represents a specific feature, e.g., a pixel can fall in the region that represents the cloud shadow and another could indicate the cloud. We then detected these two features pixel-wise by their bit values and mask out using the Google Earth Engine build-in function 'updateMask'. We then created the median composite images using the 'ee.Reducer.median()'. In this case, the Landsat images are aggregated over a given time. The cloud and shadow mask function was mapped on the median composite image to replace the masked pixels with cloudless pixels of the corresponding locations. We present the procedure in Figure \ref{four_graphs}. We got the cloudless images of our study area by extending the time used in aggregation to one year.

\subsubsection{Cloud Removal}

\textcolor{black}{Google Earth Engine (GEE) is a cloud-based geospatial processing platform that offers free access to a vast archive of satellite imagery, along with powerful computational resources and a suite of processing algorithms \cite{tamiminia2020google, kumar2018google}. Due to its scalability and ease of use, GEE has become increasingly popular for remote sensing and environmental monitoring applications \cite{zhao2021progress}.}

\textcolor{black}{In this study, we utilized GEE to preprocess Landsat imagery and remove cloud-related artifacts. Specifically, we employed the 'QA\_PIXEL' band, which encodes quality assurance information for each pixel in the form of bit-packed values. Each bit in this band corresponds to a specific condition such as cloud, cloud shadow, snow, or water presence.}

\begin{figure}
  \centering
  \medskip
  \begin{subfigure}[t]{.45\linewidth}
    \centering\includegraphics[width=1\linewidth]{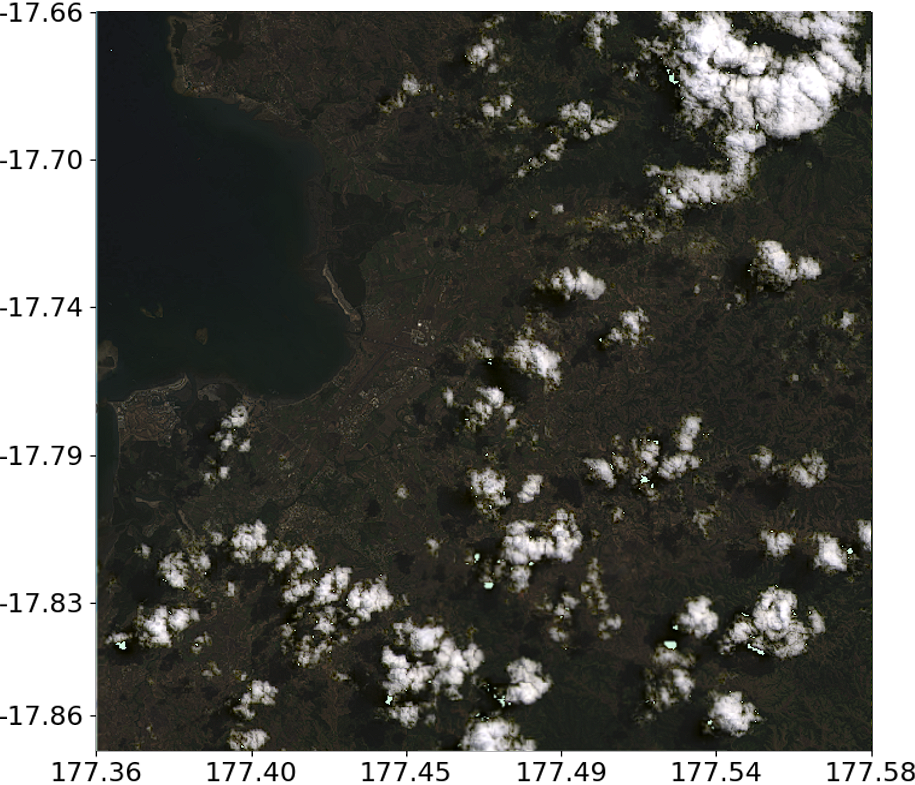}
    \caption{}
  \end{subfigure}\quad
  \begin{subfigure}[t]{.45\linewidth}
    \centering\includegraphics[width=1\linewidth]{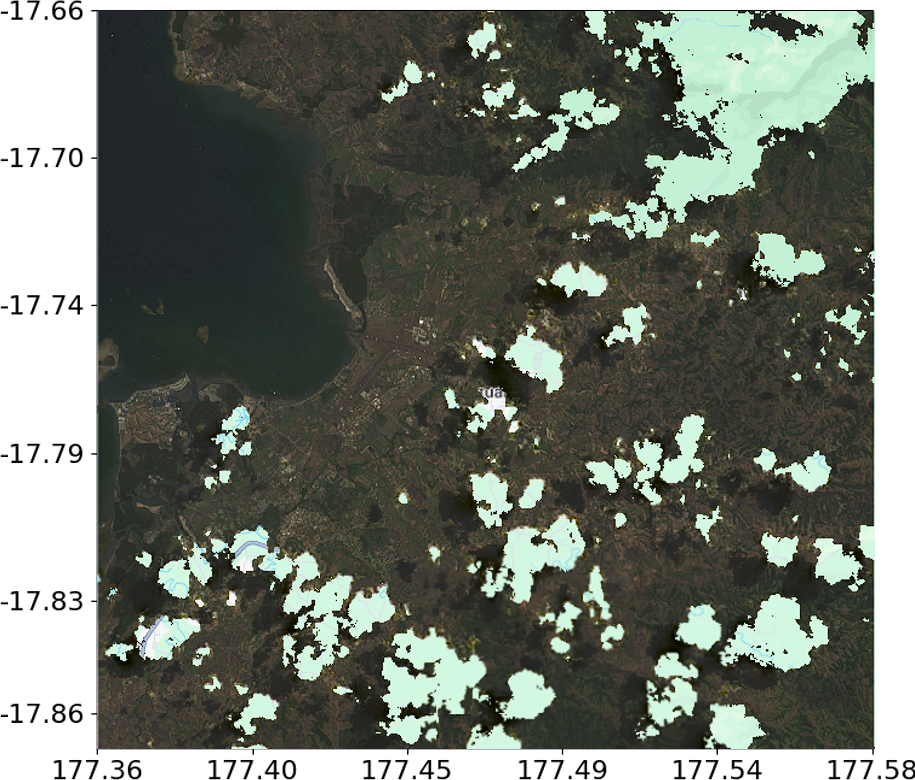}
    \caption{}
      \end{subfigure}\quad
  \begin{subfigure}[t]{.45\linewidth}
    \centering\includegraphics[width=1\linewidth]{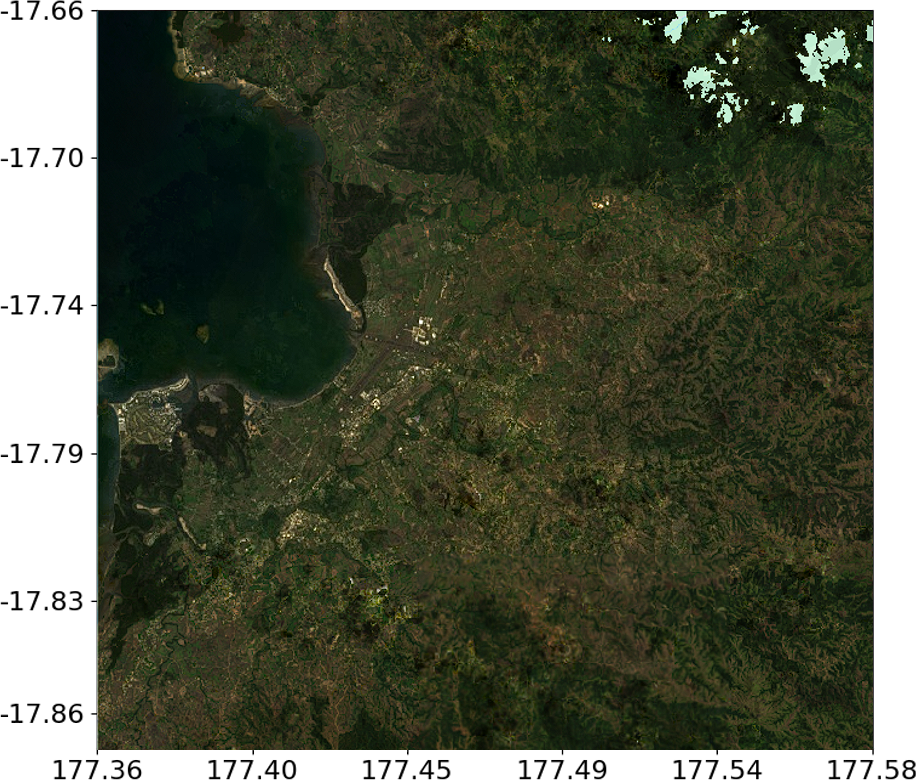}
    \caption{}
    \end{subfigure}\quad
    \begin{subfigure}[t]{.45\linewidth}
    \centering\includegraphics[width=1\linewidth]{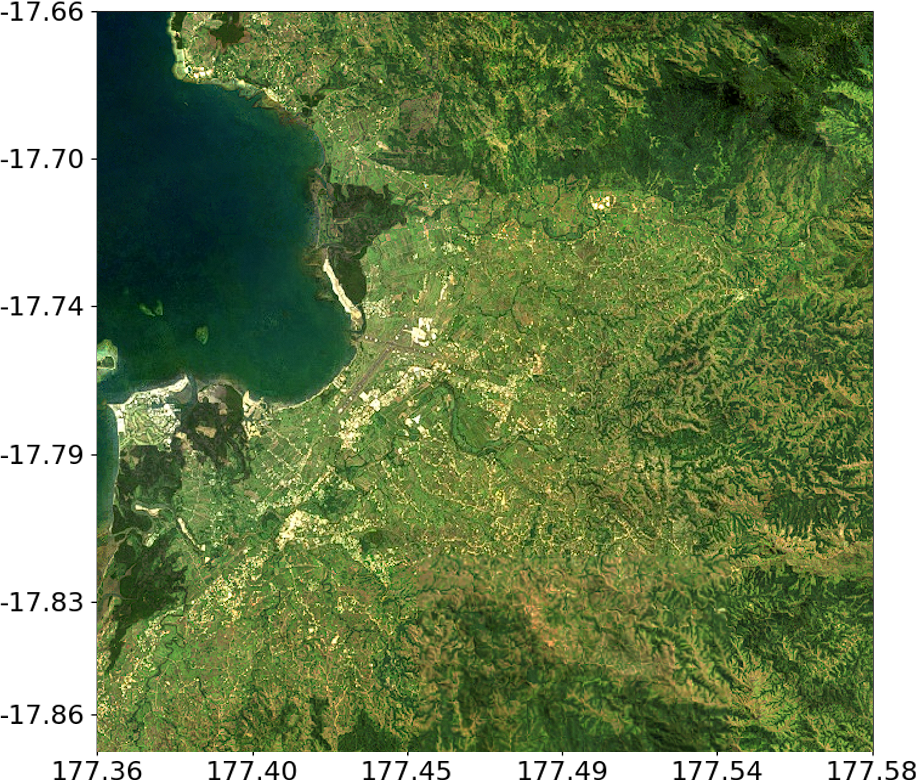}
    \caption{}
    \end{subfigure}
\caption{\textcolor{black}{Cloud and shadow removal process using Landsat 8 imagery. The sequence shows (a) the initial image with cloud cover and (b–d) the effect of applying the cloud and shadow mask on median composites over one month, three months, and one year, respectively. Longer temporal windows result in reduced cloud contamination and enhanced surface feature visibility.}}
\label{four_graphs}
\end{figure}

\textcolor{black}{To eliminate cloud and shadow contamination, we extracted the relevant bits indicating these conditions and created a pixel-wise mask using the 'updateMask()' function provided by GEE. This function allows us to mask out pixels affected by clouds and shadows while preserving clear-sky pixels.
Subsequently, we generated median composite images over a one-year temporal window using the 'ee.Reducer.median()' function. This approach aggregates all available cloud-free pixels over the specified period to create a single representative image that minimizes the impact of transient cloud cover. The entire cloud-masking and compositing workflow is illustrated in Figure~\ref{four_graphs}.}

\textcolor{black}{By extending the aggregation period to one year, we ensured the availability of sufficient cloud-free observations to produce high-quality, cloudless imagery for the study area.}

% \begin{figure}
%   \centering
%   \medskip
%   \begin{subfigure}[t]{.45\linewidth}
%     \centering\includegraphics[width=1\linewidth]{Cloud0.jpeg}
%     \caption{The first Landsat8 image collected from 2013-09-01 to 2013-10-02 that sorted by cloud cover.}
%   \end{subfigure}\quad
%   \begin{subfigure}[t]{.45\linewidth}
%     \centering\includegraphics[width=1\linewidth]{Cloud1.jpeg}
%     \caption{The cloud and shadow mask function mapped on the one-month median composite image.}
%       \end{subfigure}\quad
%   \begin{subfigure}[t]{.45\linewidth}
%     \centering\includegraphics[width=1\linewidth]{Cloud2.jpeg}
%     \caption{The cloud and shadow mask function mapped on the three-month median composite image.}
%     \end{subfigure}\quad
%     \begin{subfigure}[t]{.45\linewidth}
%     \centering\includegraphics[width=1\linewidth]{Cloud3.jpeg}
%     \caption{The cloud and shadow mask function mapped on the one-year median composite image.}
%     \end{subfigure}
% \caption{xxxxx  }
% \label{four_graphs}
% \end{figure}

\subsubsection{Index Calculations}

\textcolor{black}{In this study, we compute three widely used remote sensing indices to enhance land cover classification: the Normalized Difference Vegetation Index (NDVI), the Normalized Difference Water Index (NDWI), and the Normalized Difference Built-up Index (NDBI). These indices are calculated from multispectral bands and help discriminate between vegetation, water bodies, and built-up areas.}

\textcolor{black}{The Normalized Difference Vegetation Index (NDVI) is a well-established metric for assessing vegetation cover and health. It has been extensively used for identifying forested regions and estimating vegetation density. Studies such as \cite{tucker1979red} and \cite{pettorelli2005using} have effectively applied NDVI for land cover monitoring. NDVI values typically range from -1 to +1, where higher values indicate dense vegetation and lower values signify barren land or urban areas. It is computed using the red (R) and near-infrared (NIR) bands as follows:}

\begin{equation}
\text{NDVI} = \frac{\text{NIR} - \text{R}}{\text{NIR} + \text{R}}
\label{eq:ndvi}
\end{equation}

\textcolor{black}{To distinguish water bodies, the Normalized Difference Water Index (NDWI) was introduced by McFeeters in 1996 \cite{mcfeeters1996use}. It utilizes the NIR and short-wave infrared (SWIR) bands, providing good contrast between water and other land types:}

\begin{equation}
\text{NDWI} = \frac{\text{G} - \text{NIR}}{\text{G} + \text{NIR}}
\label{eq:ndwi}
\end{equation}

\textcolor{black}{However, NDWI may sometimes fail to clearly separate water from vegetation. To address this, Xu (2006) proposed the Modified NDWI (MNDWI) \cite{xu2006modification}, which replaces NIR with the green (G) band to improve water body detection. MNDWI has shown improved performance in complex environments, especially where water and vegetation are interwoven \cite{szabo2016specific}. It is calculated as:}

\begin{equation}
\text{MNDWI} = \frac{\text{G} - \text{SWIR}}{\text{G} + \text{SWIR}}
\label{eq:mndwi}
\end{equation}

\textcolor{black}{The Normalized Difference Built-up Index (NDBI) \cite{zha2003use} is used to highlight built-up areas and mitigate classification errors caused by topographic effects. It is derived using the SWIR and NIR bands as shown below}

\begin{equation}
\text{NDBI} = \frac{\text{SWIR} - \text{NIR}}{\text{SWIR} + \text{NIR}}
\label{eq:ndbi}
\end{equation}

\textcolor{black}{Together, these indices play a crucial role in improving the accuracy of land cover classification by enhancing the spectral separability of various land cover types.}

\subsubsection{Classification scheme}

This study adopts the European Union Land Use/Cover Area Frame Survey (LUCAS) \cite{gallego2010european} classification scheme. The land cover in the scheme has been categorized into eight classes and 29 subclasses. Due to the medium resolution of the data, the classification scheme in our study focuses on the major classes  which we modified based on the study area. We combined the cropland and grassland into one class due to their similar spectral signatures and close growth locations. The shrubs exhibit sparse coverage, where grass and forest can be observed within 30 meters and separate shrubs into one class can lead to spectral confusion. Therefore, we removed the shrubland   from the scheme and added the coastal area as a separate class. As a result, we use a classification scheme with seven classes  including, Urban Areas, Grass/Agricultural Land, Forest, Bare Soil, Water Bodies, Coastal Areas and Wetland.  However, we discovered that each land cover type in the scheme encompasses different applications and to mitigate the misleading arising from ambiguous delineation among types, we combined artificial grasslands like the football pitch and mowed lawns into grassland type. We classified the roads built with sand and the cropland without planted vegetation as bare soil. The impervious surfaces and the construction areas are classified as urban areas. The wetlands are saturated with water bodies found along the coast, in the mangrove forests and estuaries \cite{guo2017review}, where the marshes are also a type of wetland \cite{kesikoglu2019performance}.

%xxxx

\subsection{Data collection and labelling} 

The supervised machine learning algorithms require a labelled training dataset. In remote sensing applications, a major challenge is to create a labelled training dataset. The scarcity of labelled data in remote sensing applications is well known \cite{dutta2023remote}, and there have also been approaches to automatically label them \cite{huang2015automatic}. However, automatic labelling of data would need expert knowledge and some manual labelled data, which differed from application to application. 

In our study, we manually generalled ground-truth label by  by observation and   analysis of Google Earth Engine high-resolution historical images,
%where we located xxxx xxx and xxx for labelling.
\textcolor{black}{Distinct land cover types were identified and labelled based on observable characteristics such as spatial texture, tone, shape, and contextual cues within the imagery.}

Each point was randomly picked, and we collected samples from dispersed locations, ensuring diversity. Knowing that there is no unified standard for the sample size employed in the land cover classification, the experiment has been done to find a suitable sample size for our study, and the results are discussed in the later chapter.

\subsection{Chip Generation} 
The image chips are the small sub-images generated from the original image, which helps us process large datasets and train the model. The satellite image of our study area has a size of 780*818 pixels and 10 channels. The 'chopped' images can save computation time and memory in processing. The adjacent pixels are divided into one image chip, allowing algorithms like neural networks to learn from the spatial context. 
We defined 9*9 pixels as our chip size and allowed padding of y size/2 and x size/2 to our original image. A total of 638040 chips have been generated to allow algorithms to learn and classify each pixel based on its spatial context. The samples used in training and testing were generated by associating image chips with their corresponding ground truth labels.

\subsection{Machine learning methods}

\subsubsection{K-means} 

%. K-means is a commonly used unsupervised learning algorithm in land classification.

K-means is an unsupervised learning method that has been widely used for classifying land covers in many studies \cite{ usman2013satellite,lv2019novel,abbas2016k}.
K-means assign objects with similar characteristics into one cluster \cite{usman2013satellite}, and the clusters are disjoint \cite{likas2003global}. The procedure requires specifying k points as the initial centroids of the clusters and assigning objects to clusters based on their distance to the centroids \cite{lloyd1982least}. In order to minimize the errors in clustering, the algorithm is iterated to minimize the squared error objective function through repeated calculation of the centroid's position and re-assigning the members \cite{sinaga2020unsupervised}. The function is shown as follow: $$ J(z,A)= \sum_{i=1}^{n}\sum_{k=1}^{c} \left \| x_{i}-a_{k} \right \|^{2} $$
Where the $a_{k}$ are the centroids of the clusters, and $x_{i}$ are the observations. The group assignations are finally fixed until the position of the centroids is fixed. 

  K-means has the advantage due to its convenience. It can be applied to unlabeled data and has a relatively short calculation time due to its linear time complexity, making it suitable for classifying a large dataset.
The disadvantage of using K-means in land cover classification is that the method is limited in preserving spatial context. K-means grouping individual pixels based on their characteristic makes classifying mixed land cover inappropriate. Also, the algorithm is sensitive to the initial number of centroids k. The results can be varied with different selections of k  value.

\subsubsection{Random Forest} 

Random Forest is an ensemble learning  method consisting of a combination of random forest trees. The procedure involves Bootstrap to create random subsets of original training data, and the trees are grown on the diverse subsets \cite{breiman1996bagging}.
The random forest can be expressed using: $$\{ h(\textbf{x},\Theta_{k}), k = 1,.. \}$$ where $k$ is the number of trees, $\textbf{x}$ is the input vector, and $\Theta_{k}$ are the independent identically distributed random vectors \cite{breiman2001random}. The output-generating process can be expressed as: $$\widehat{\textup{C}}_{rf}^{B} = majorityvote\left \{ \widehat{\textup{C}}_{b}(\textbf{x})\right \}_{1}^{B}$$ where $\widehat{\textup{C}}_{b}(\textbf{x})$ is the prediction generated by $b_{th}$ tree \cite{rodriguez2012assessment}. The final prediction combined the decision from every single tree by majority vote.

\subsubsection{Multilayer Perceptron}
 
The multilayer perceptron (MLP) also known as a simple neural network, is a powerful tool in analysing remote sensing data \cite{zhu2017deep, talukdar2020modeling}. The MLP consists of input layer, hidden layer and output layer.  The nodes among layers are connected with scaled nonlinear output by weight, the model adjusts the approximation function by superpositioning nonlinear transfer functions. The connection between nodes can be expressed as: $$ y=\varphi\sum_{i=1}^{n}w_{i}x_{i}+b$$ w indicates the weights, x are the input objects, b is the bias, and $\varphi$ is the non-linear activation function \cite{jamali2019evaluation}.  The MLP maps output from input by featuring  nonlinear activation functions via gradient-based training algorithms such as backpropagation\cite{gardner1998artificial}.

\subsubsection{Convolutional Neural Network} 
CNN is one of the popular methods used in image processing, which consists of the input layer, the convolutional layer, the pooling layer, the fully connected layer and the output layer. One advantage of CNN is its ability to learn spatial features from various and successive data transformations between convolutions and poolings \cite{kattenborn2021review}. When dealing with digital images, the input of CNN is the two-dimensional (2D) grids comprising digital values. The convolution layers extract salient features by applying filters at each position of the image \cite{yamashita2018convolutional}. The pooling operation reduces the dimension of the feature images from convolution layers. The multi-dimensioned data are then flattened into a vector and passed to the fully-connected layer. Finally, the 'Softmax' function assigns probabilities to each class, where the class with the highest probability becomes the output. The Convolutional Neural Network is trained to eliminate the difference between prediction and actual value through backpropagation and gradient descent \cite{zubair2020parameter}. The disadvantage is that CNN requires a large sample size to train the model \cite{han2018new}. Also, the training can be time-consuming since the convolution process is slow, especially for a deep network \cite{jogin2018feature}.

\subsection{Framework}

% In this study, we present a framework that utilises a combination of supervised and unsupervised machine learning models  including, k-means clustering, random forests, multilayer perception, and CNNs  in stages highlighted in Figure \ref{fig:framework}.
% The goal of the framework is to provide a study of the LULCC in the selected region (Figure 1) of Nadi, Fiji from 2013 onwards.

% In Step 1 of the framework in Figure \ref{fig:framework}, we begin by obtaining and processing the remote sensing data for the selected region. 

% In Step 2, we take a subset of the training and manually create labels. This would be also seen as a "human in the loop" approach, where human expertise is needed for automating a machine learning framework \cite{wu2022survey}.

%  %However, rare studies utilize the Google Earth Engine platform for implementation. 

%  In Step 4, we train three classification models, that includes multilayer perception, CNN and random forests and evaluate their performance. 

%  In Step 5, we use   K-means clustering  to show its applicability in classifying multiple land covers on the Google Earth Engine platform, rather than compare it with other supervised learning algorithms. The same data is  used for all three supervised learning models and compared with k-means clustering which does not use labelled data. This is to provide a comparison and also evaluate how much effect the labelled data has on the framework.

%   In Step 6, we take the best model from Step 4 and 5, and use for LULCC study for the study area in Nadi and study changes since 2013. 
\begin{figure*}[!htb]
    \centering
    \includegraphics[width=1\linewidth]{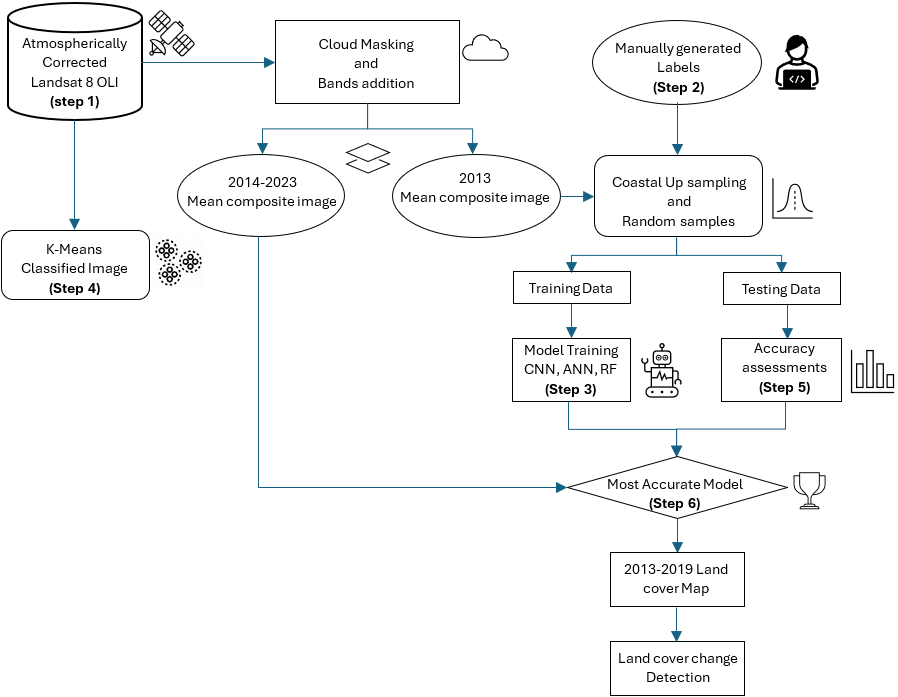}
    \caption{\textcolor{black}{Framework for land use and land cover classification. It includes data normalisation, manual labelling with coastal upsampling, supervised (CNN, RF, ANN) and unsupervised (K-Means clustering) training, model evaluation, and selection of the best model for future classification.}}

    %give link to fig so I can edit
    \label{framework}
\end{figure*}

\textcolor{black}{In this study, we propose a comprehensive framework that integrates both supervised and unsupervised machine learning techniques—including k-means clustering, random forests, multilayer perceptron (MLP), and  CNN, as illustrated in Figure~\ref{framework}. The framework aims to analyze land use and land cover change (LULCC) in the selected region of Nadi, Fiji, from 2013 onwards (Figure ~\ref{fiji_topo}).}

\textcolor{black}{In \textbf{Step 1}, the process begins with the acquisition and preprocessing of remote sensing data for the study area. This involves cloud masking, compositing, and extracting relevant spectral features required for classification.}

\textcolor{black}{In \textbf{Step 2}, a subset of the dataset is selected for manual labelling to create a ground-truth training dataset. This manual annotation process embodies a ``human-in-the-loop'' approach, where expert knowledge guides the development of an effective machine learning workflow \cite{wu2022survey}.}

\textcolor{black}{In \textbf{Step 3}, three supervised classification models—MLP, CNN, and random forests—are trained and evaluated on the prepared dataset. Each model’s performance is assessed using standard accuracy metrics to identify its suitability for LULCC tasks.}

\textcolor{black}{In \textbf{Step 4}, unsupervised classification is performed using k-means clustering. This step is designed not to directly compete with the supervised methods but to demonstrate the applicability of unsupervised techniques for multi-class land cover classification within the Google Earth Engine environment.}

\textcolor{black}{In \textbf{Step 5}, the performance of all models is compared to determine the most effective approach for classifying the region’s land cover. This comparison highlights the added value of labelled datasets and the role of supervised learning in achieving higher accuracy.}

\textcolor{black}{Finally, in \textbf{Step 6}, the best-performing model is employed to conduct a temporal analysis of LULCC in the Nadi region. The model is applied to multi-year satellite imagery to detect and quantify changes in land cover since 2013.}

\subsubsection{Model metrics}

  There are four indices commonly used in assessing classification predictions, including true positives (TP),  true negatives (TN),  false positives(FP) and false negatives(FN). These four indices play a significant role as they usually provide details in the classification. These indices are usually plotted on a confusion matrix, which allows people to readily locate the class that has been misclassified. The change detection in this study is delivered by comparing the proportion of the target class in the total study area, and this makes identifying the class that has been overestimated or underestimated crucial for our study, as it directly affects the proportion of the class that has been misclassified.  

  Additionally, the misclassifications are not always caused by the model, and there are various reasons behind the problem. Analysing the recurred misclassifications allows us to make improvements in the methods. For example, if the Grass/Agricultural Land and Forest are misclassified in the experiment, we may consider selecting non-adjacent samples when collecting; If the water bodies and vegetation are misclassified, we might consider adding water and vegetation indices as input.

  The accuracy score, precision score, recall and F1 score were calculated in this study to assess the classification performance of the models.The formulas of each are as follows:
$$Accuracy = \frac{TP+TN}{TP+TN+FP+FN}$$
$$Precision = \frac{TP}{TP+FP}$$
$$Recall = \frac{TP}{TP+FN}$$
$$F1  = \frac{2*TP}{2*TP+FP+FN}$$

 In order to compare the learning process, we have plotted learning curves for both ANN and CNN models. The learning curves are used to detect the fitness of the model. The model indicates under-fitting when the loss stays high and does not decrease with an increase in epochs. The model indicates that it performs well in the training dataset and worse in the unseen datasets. In the learning curve, it happens when the training loss is much lower than the test loss.

% \subsubsection{Technical setup}

% In this study, we used 70\% of the samples for training and 30\% for tests. The model was trained yearly, and the training datasets were generated each year. 

\subsubsection{Technical setup}

\textcolor{black}{In this study, we employed 10-fold cross-validation for model evaluation. The dataset was randomly partitioned into 10 equal subsets, where in each iteration one subset was used for testing and the remaining nine for training. The process was repeated 10 times, and the average performance across all folds was reported. The model achieving the best accuracy among the folds was selected as the final model.}

%%%%%%%%%%%%%%%%%%%%%%%%%%%%%%%%%%%%%%%%%%%%%%%%%%%%%%%%%%%%%%%%%%%%
\section{Results}
\label{sec:sample1}
%%%%%%%%%%%%%%%%%%%%%%%%%%%%%%%%%%%%%%%%%%%%%%%%%%%%%%%%%%%%%%%%%%%%
\subsection{Selection of sample size}

Taking random samples is a commonly used method when predicting a large population, and choosing an appropriate sample size helps generate accurate results. A small sample size can not represent the whole population since it does not cover the diversity of the population. In some cases, even many outliers or extreme values are encountered. In the land cover classification, a selection of small samples can not encompass the characteristics of land cover types. However, generating a large sample can be time-consuming. The current studies lack a unified standard for the number of samples employed in the land cover classification. Vasuti et al. \cite{vasuki2019spatial} selected 141 samples to classify vegetation, non-vegetation and water in the Darling Range, Western Australia, and the sample size was approximately 0.03\% of their total study area. Tu et al. \cite{Tu2020} selected 6455 samples to classify eight land cover types in the Guangdong Province, China. Both the statistics and machine learning fields provided suggestions on the sample size.

\begin{figure*}[!htb]
  \centering
  \medskip
  \begin{subfigure}[t]{.45\linewidth}
    \centering\includegraphics[width=0.8\linewidth]{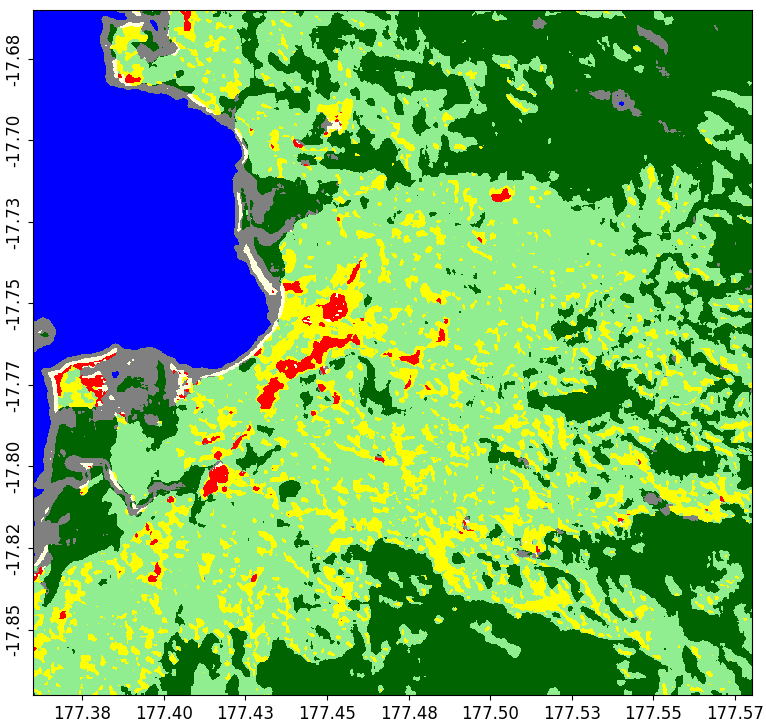}
    \caption{490 samples}
  \end{subfigure}\quad
  \begin{subfigure}[t]{.45\linewidth}
    \centering\includegraphics[width=0.8\linewidth]{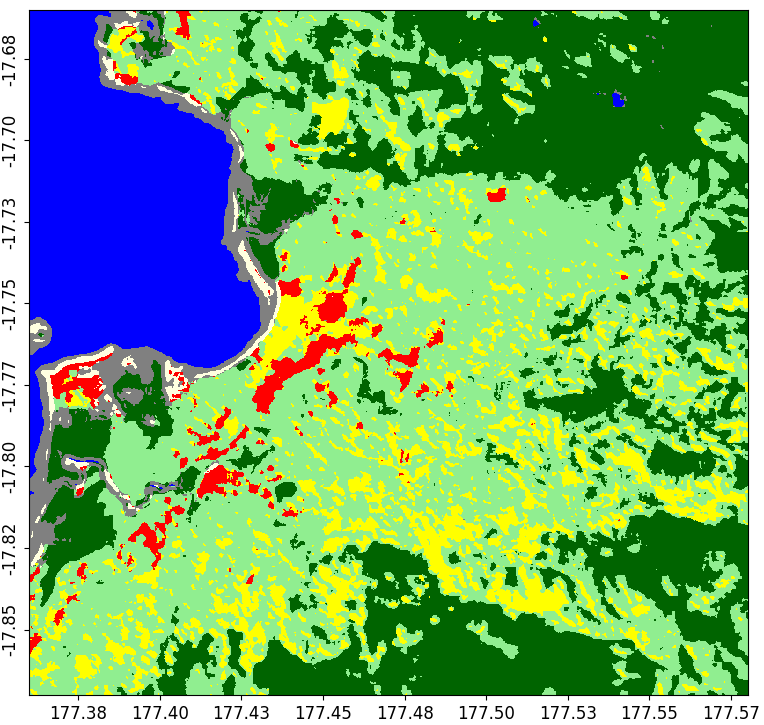}
    \caption{700 samples}
      \end{subfigure}\quad
  \begin{subfigure}[t]{.45\linewidth}
    \centering\includegraphics[width=0.8\linewidth]{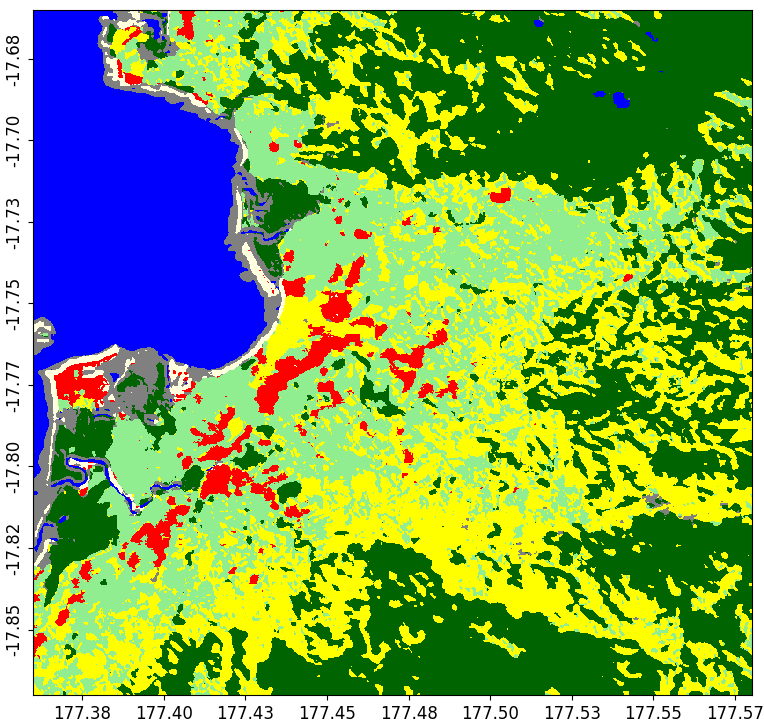}
    \caption{1050 samples}
    \end{subfigure}\quad
    \begin{subfigure}[t]{.45\linewidth}
    \centering\includegraphics[width=0.8\linewidth]{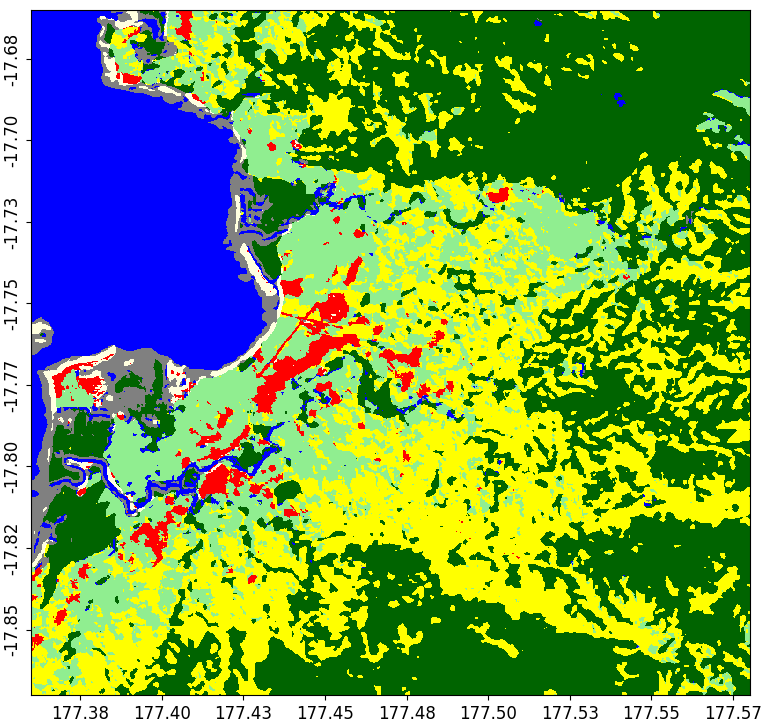}
    \caption{1400 samples}
    \end{subfigure}
    \begin{subfigure}[t]{.45\linewidth}
    \centering\includegraphics[width=0.985\linewidth]{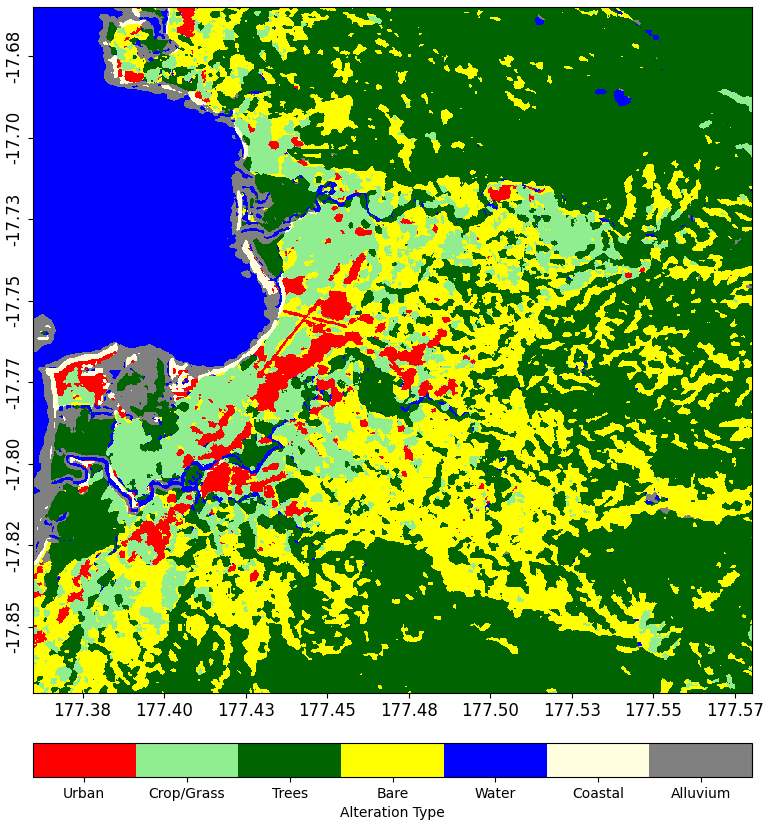}
    \caption{1750 samples}
    \end{subfigure}
\caption{a) The land cover map plot using different sample sizes}
\label{land cover_sample zise}
\end{figure*}

  In statistics, the sample size used to classify 638040 pixels is 384 with a 95\% confidence level and 5\% confidence interval. In machine learning, the "rule of thumb" suggests a minimum amount of reference data should be $m \geq 10 n \cdot C$, where m is the sample size, n represents the input types, and C indicates the output classes. In this study, the minimum amount of sample size should be 490. Knowing that noises in the samples are likely, possibly caused by unavoidable human error in labelling. CNN has been used to examine the impact of sample size on the results since neural networks are robust to training data heterogeneity \cite{paola1995review}. We used a total of 490 samples (70 samples for each class), 700 samples (100 samples for each class), 1050  samples (150 samples for each class), 1400 samples (200 samples for each class) and 1750 samples (250 samples for each class) in the experiments to find a suitable sample size. The land cover maps plotted using each sample size are shown in the Figure \ref{land cover_sample zise}. 

% Cover map with different sample size start here
%-------------------------------------

\begin{table}[!htb]
\caption{The assessment score of the CNN model using different sample sizes}
\scalebox{1}{
\begin{tabular}{||c  c c c c||} 
 \hline
 sample size & accuracy & precision & recall & f1 score \\
 \hline\hline 
  490  & 0.9659 & 0.9677 & 0.9659 & 0.9657\\ 
  
  700  & 0.9523 & 0.9530 & 0.9523 & 0.9521\\
  
  1050 & 0.9809 & 0.9812 & 0.9809 & 0.9809\\

  1400 & 0.9857 & 0.9881 & 0.9782 & 0.9825\\

  1750 & 0.9943 & 0.9935 & 0.9964 & 0.9949\\
 \hline
\end{tabular}}
\label{assessment score_sample sizes}
\end{table}

%---------------------------------------
\vspace{5mm}
 \noindent The accuracy scores (table \ref{assessment score_sample sizes}) are high for each of the sample sizes. The training loss and testing loss are close, and they both converged quickly and became stable. However, the land cover maps are varied, using different sample sizes. Due to the high accuracy, we have first considered the 'Accuracy paradox', which happens when the model is biased but has a high accuracy score. The precision, recall and F1 score are then checked, ensuring the model is unbiased. One possible reason is that the minimum sample size required in theory can not capture the diversity of the urban area, since the colour of the house roof in our study area is varied. 

\vspace{5mm}
%  \noindent  The land cover map from each experiment is visually compared with the ground truth image, we then found that the urban areas are effectively captured using 1750 samples. This result coincides with Colditz's \cite{paola1995review} study, where he claimed that the samples should be approximately 0.25\% of the total pixels. The coastline is narrow in our study, making collecting samples difficult. We collected 91 samples of coastal regions yearly and up-sampled them to balance the dataset. 

% \begin{table}[!htb]
% \centering 
% \caption{The final selection of samples in each land cover type.}
% %\scalebox{0.63}{
% \begin{tabular}{c c c c} 
%  \hline
%  Types & Training samples &Test samples &  Proportion   \\
%  \hline\hline 
%   Urban Areas & 175 & 75 & 0.038\% \\ 
%   Grass/Agricultural Land & 175 & 75 & 0.038\%\\ 
%   Forest & 175 & 75 & 0.038\% \\ 
%   Bare Soil & 175 & 75 & 0.038\% \\ 
%   Water Bodies & 175 & 75 & 0.038\% \\  
%   Coastal Areas & 175 & 75 & 0.038\% \\ 
%   Wetland & 175 & 75 & 0.038\% \\ 
%   \hline\hline
%  Total & 1225 & 525 & 0.26\% \\ 
%  \hline
% \end{tabular}%}
% \end{table}

\noindent
\textcolor{black}{The final selection of samples for each land cover type included 175 training samples and 75 test samples for each of the seven classes. This allocation represents approximately 0.038\% of the total study area for each class. In total, 1225 training samples and 525 test samples were used, covering 0.26\% of the total pixels. Visual comparison of the land cover maps from each experiment with the ground truth image revealed that urban areas were most effectively captured using 1750 samples. This finding aligns with Colditz’s study~\cite{paola1995review}, which suggests that sample sizes should be approximately 0.25\% of the total pixels. Due to the narrow coastal regions in the study area, collecting sufficient samples was challenging; hence, 91 coastal samples were collected annually and up-sampled to balance the dataset.}

%------------------------------------%

\subsection{Model construction}
\noindent \textbf{K-means:} We run the K-means algorithm on the Google Earth Engine using the package 'ee.Clusterer', this algorithm is based on the machine learning algorithm software Weka in Java. We selected the region we wanted by clipping the image with a pre-defined data frame. When the remotely sensed satellite images are passed into the algorithm, the clusters represent different land cover types. The number of clusters in this study is seven. Each pixel in the image is labelled with a given index representing the clusters, and the pixels with the same cluster index are categorised into one land cover type.
\vspace{5mm}

\textcolor{black}{\noindent \textbf{Random Forest:} The hyperparameters of the random forest algorithm were selected using a five-stage grid search, including ‘max depth’, ‘max features’, ‘min samples leaf’, ‘min samples split’, and ‘n estimators’. The values of each parameter used in the experiment are provided in the table. Model performance was evaluated using 10-fold cross-validation.}

\vspace{5mm}
\textcolor{black}{\noindent \textbf{Artificial Neural Networks:} The model inputs were 1225 flattened vectors derived from image chips of size $9 \times 9 \times 10$. Hyperparameters were optimized using grid search, including the number of layers, neurons per layer, optimizer, and activation function. Training was monitored on accuracy, with early stopping (patience = 15) to prevent overfitting. The final ANN architecture consisted of one hidden layer with 32 neurons, the ‘Adam’ optimizer, ReLU activation, and categorical cross-entropy as the loss function. Training was conducted for a maximum of 150 epochs with a batch size of 32. Model evaluation was based on 10-fold cross-validation, and the best-performing fold was taken as the final result.}

\vspace{5mm}
\textcolor{black}{\noindent \textbf{Convolutional Neural Networks:} The model inputs were 1225 image chips of size $9 \times 9$ pixels with ten channels. The CNN architecture consisted of three convolutional layers with 32, 48, and 64 kernels of size $1 \times 1$, followed by dropout layers (0.25 and 0.5) to mitigate overfitting. The output was flattened and passed through a fully connected layer. ReLU activation and the Adam optimizer were used, with sparse categorical cross-entropy as the loss function. Training was performed for up to 150 epochs with early stopping (patience = 15). Model performance was assessed using 10-fold cross-validation, with the best-performing fold taken as the final result.}

%------------------------------%
\subsection{Score comparison}

In order to choose the model with the best classification performance, we have compared the accuracy scores of three supervised learning models, which were delivered by dividing the number of correctly classified samples by the total test samples. The classification accuracies for all three models were high. ANN had an accuracy of 98.29\%, and CNN had an accuracy of 99.43\%. 
 The confusion matrices are shown in the Figure. In general, all three models performed well in classifying the Coastal Areas and wetlands while making mistakes in classifying the  Grass/Agricultural Land, Forest and Water Body. We especially cared about the performance of the models in classifying the urban areas. The distribution of houses is sparse in the eastern side of our study area, close to the mountain areas, making distinguishing between bare land and urban area difficult due to their similar spectral reflectance. CNN performed best among the three models. This could be a good sign for us since an accurate classification of the urban areas is essential for change detection. 

\begin{table}[!htb]
\centering 
\caption{Accuracy of supervised learning models over 10-fold cross-validation.}
\scalebox{1}{
\begin{tabular}{||c c c c||} 
 \hline
 Fold & CNN & RF & ANN \\ 
 \hline\hline
1  & 0.9429 & 0.9714 & 0.9600\\ 
2  & 0.9657 & 0.9600 & 0.9543\\ 
3  & 0.9714 & 0.9771 & 0.9657\\ 
4  & \fcolorbox{black}{yellow}{\textbf{0.9943}} & 0.9771 & 0.9886\\ 
5  & 0.9829 & \fcolorbox{black}{yellow}{\textbf{0.9829}} & \fcolorbox{black}{yellow}{\textbf{0.9829}}\\ 
6  & 0.9657 & 0.9771 & 0.9714\\ 
7  & 0.9714 & 0.9543 & 0.9600\\ 
8  & 0.9771 & 0.9657 & 0.9771\\ 
9  & 0.9771 & 0.9714 & 0.9771\\ 
10  & 0.9486 & 0.9486 & 0.9771\\
\hline\hline
Mean & 0.9697 & 0.9686 & 0.9714\\
Std & 0.0002090 & 0.0001123 & 0.0105374\\
\hline
\end{tabular}}
\label{tab:average_scores_30_runs}
\end{table}

\begin{table}[!htb]
\centering 
\caption{The assessment score of supervised learning models}
\scalebox{1}{
\begin{tabular}{|| c c c c c||} 
 \hline
 model & accuracy & precision & recall & f1 score \\ [1ex] 
 \hline\hline
RF  & 0.9829 & 0.9830 & 0.9849 & 0.9836\\[2ex]
ANN  & 0.9886 & 0.9870 & 0.9893 & 0.9879\\[2ex]
CNN & 0.9943 & 0.9935 & 0.9964 & 0.9949\\[2ex]
 \hline
\end{tabular}}
\label{assessment model}
\end{table}

% Compare Confusion matrix
\begin{figure}[!htb]
  \centering
  \medskip
  \begin{subfigure}[t]{0.45\linewidth}
    \centering\includegraphics[width=1\linewidth]{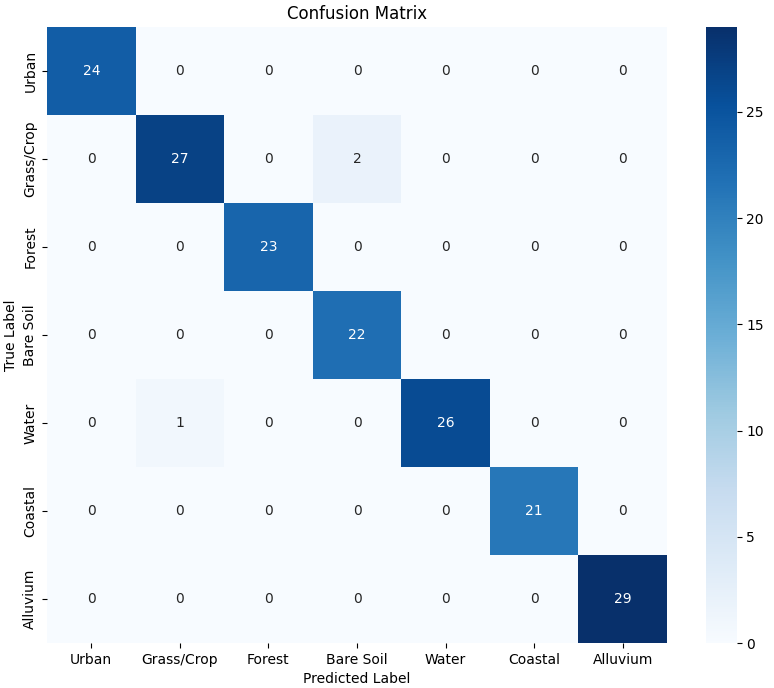}
    \caption{RF}
    \end{subfigure}
    \begin{subfigure}[t]{0.45\linewidth}
    \centering\includegraphics[width=1\linewidth]{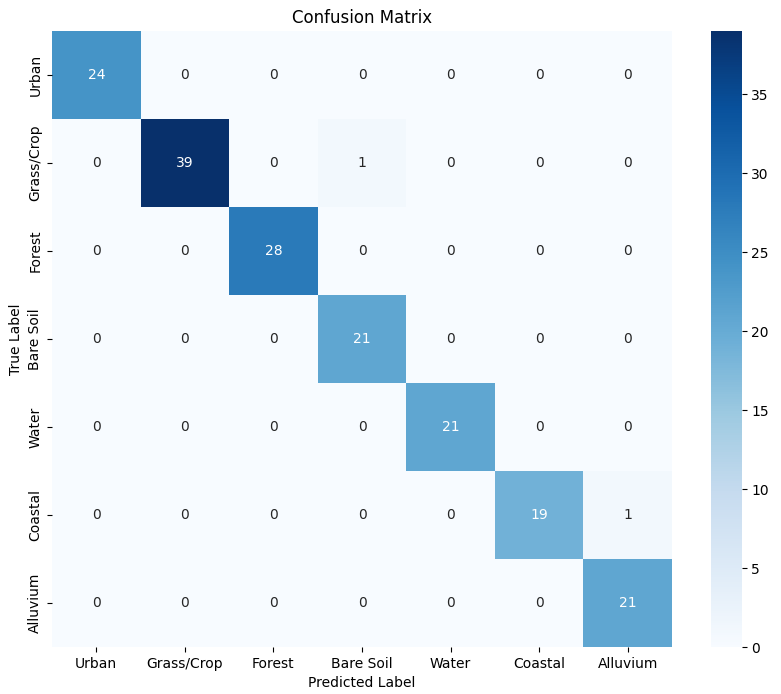}
    \caption{ANN}
    \end{subfigure}
    \begin{subfigure}[t]{0.9\linewidth}
    \centering\includegraphics[width=1\linewidth]{confusion_matrix_rf.png}
    \caption{CNN}
    \end{subfigure}    
\caption{The confusion matrices of supervised learning models}
\label{Confusion matrix}
\end{figure}

% Compare ROC_AUC
\begin{figure}[!htb]
  \centering
  \medskip
    \begin{subfigure}[t]{0.45\linewidth}
    \centering\includegraphics[width=1\linewidth]{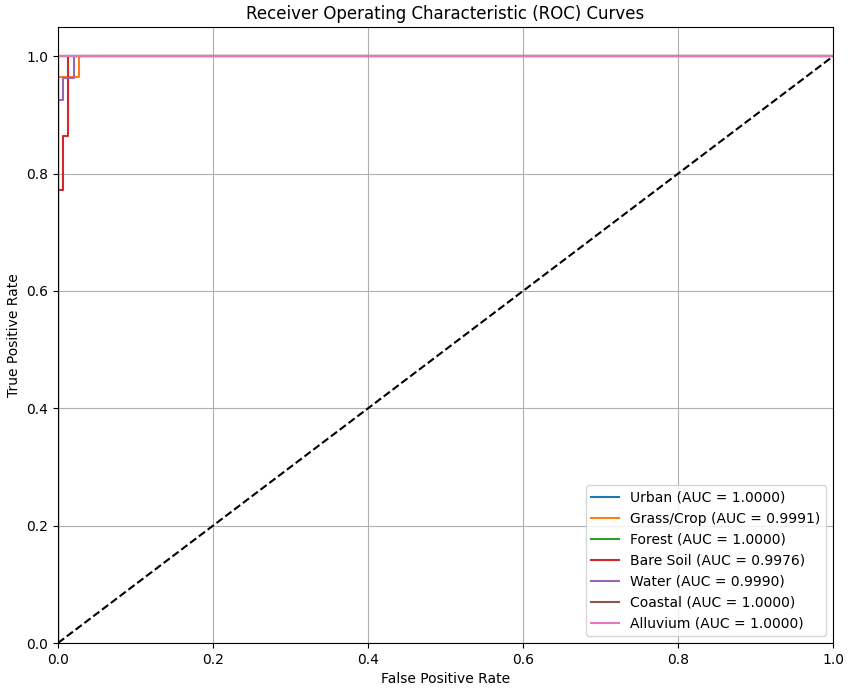}\\
    \caption{RF}
    \end{subfigure}
    \begin{subfigure}[t]{0.45\linewidth}
    \centering\includegraphics[width=1\linewidth]{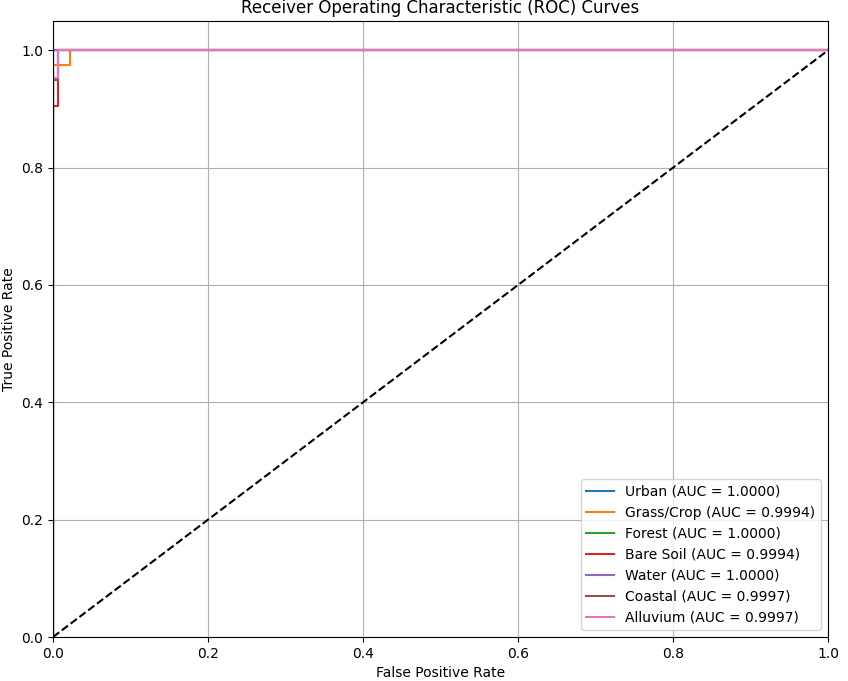}\\
    \caption{ANN}
    \end{subfigure}
    \begin{subfigure}[t]{1\linewidth}
    \centering\includegraphics[width=1\linewidth]{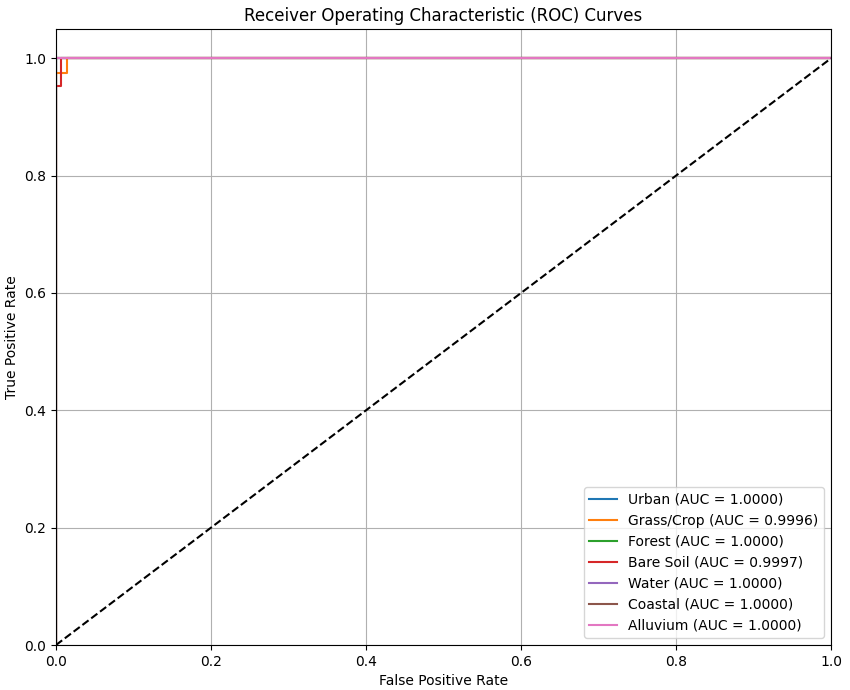}
    \caption{ CNN}
    \end{subfigure}    
\caption{The ROC AUC curves of supervised learning models}
\label{ROC_AUC}
\end{figure}

\subsection{Map comparison}
\subsubsection{Unsupervised learning method}
% \begin{figure}[!htb]
%     \centering
%     \includegraphics[width=7cm, height=7cm]{K-Means.png}
%     \caption{The land cover map plot using K-means}
%     \label{K-means_map}
% \end{figure}

The k-means model overestimated the urban areas. We found that the misclassification in the eastern mountain regions was conspicuous since the algorithm recognised certain burnt-bare soils and red soil as urban areas. The dense forests were assigned a white colour, while sparser forests were assigned a dark green colour, and the tributaries were classified as dense forests.

\subsubsection{Supervised learning method}

\begin{figure}[!htb]
  \centering
  \medskip
    \begin{subfigure}[t]{.45\linewidth}  \centering\includegraphics[width=1\linewidth]{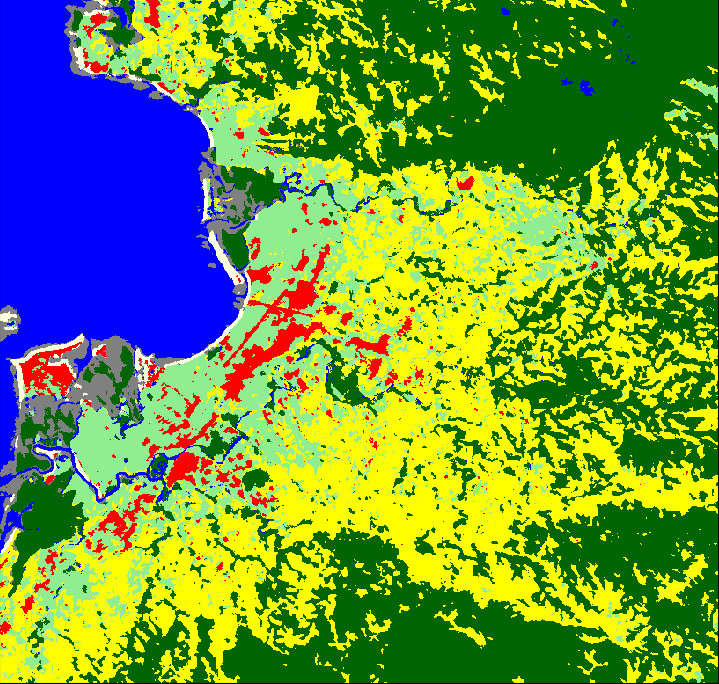}
    \caption{ RF}
    \end{subfigure}
    \begin{subfigure}[t]{.45\linewidth}
    \centering\includegraphics[width=1\linewidth]{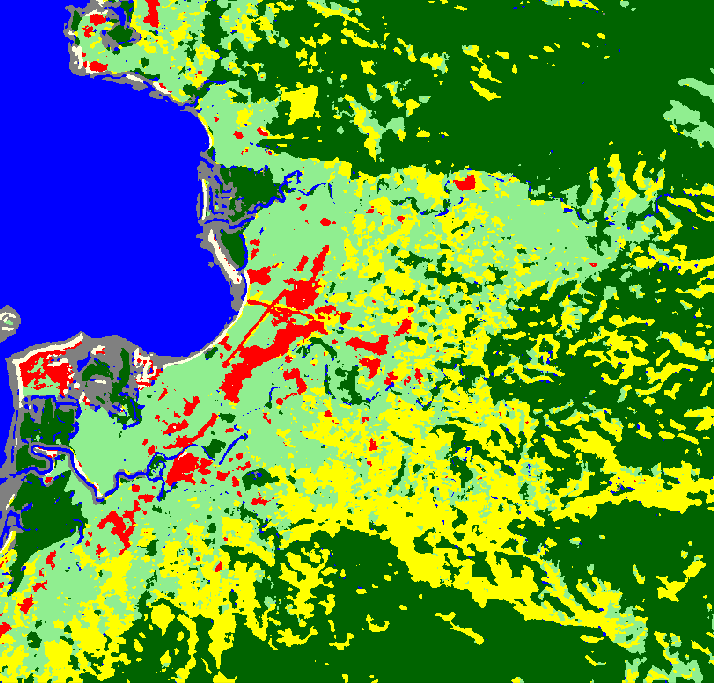}
    \caption{ANN}
    \end{subfigure} 
    \begin{subfigure}[t]{.45\linewidth}  \centering\includegraphics[width=1\linewidth]{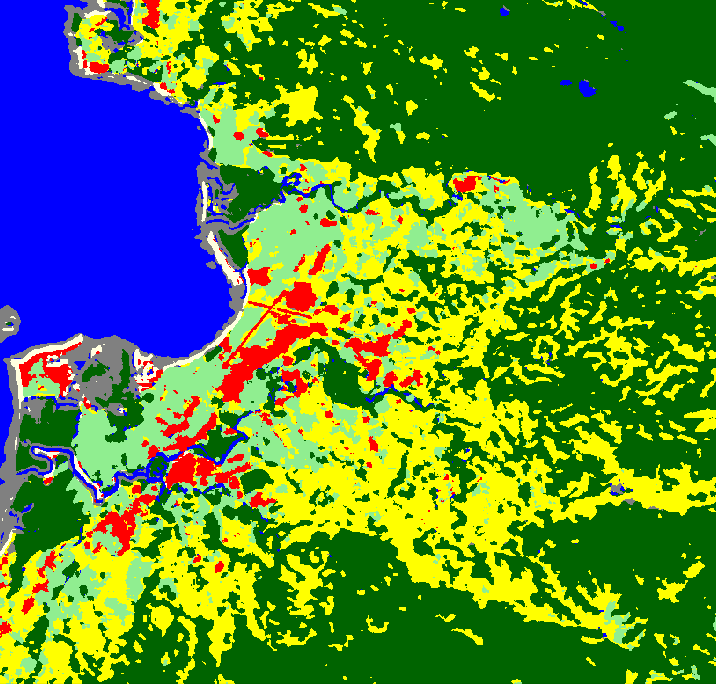}
    \caption{CNN}
    \end{subfigure}
    \begin{subfigure}[t]{.425\linewidth}
    \centering\includegraphics[width=1\linewidth]{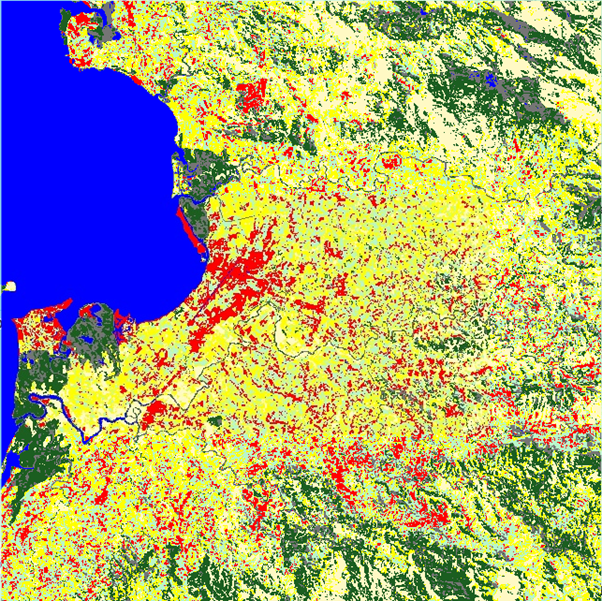}
    \caption{K-Means}
    \end{subfigure}
    \begin{subfigure}[t]{1\linewidth}
    \centering\includegraphics[width=0.9\linewidth]{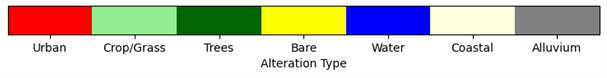}
    \end{subfigure}

\caption{Classified maps comparison }
\label{}
\end{figure}

The assessment scores and the performance in classifying the urban area indicate that CNN is the best model for classifying the land covers in our study area. Thus, we chose CNN model for classifying the land cover types of the rest years' images.

%\section{Results}
%\label{sec:sample1}
\subsection{Landcover Maps}

We found that the classification scheme was appropriate. Overall, the introduced classes described the land covers in our study area. The algorithms performed well in separating the bare soil and grass/Agricultural Land. This can be seen intuitively from the classified results of the rural area and the mountain area. The algorithm showed its capability in classifying large-scale grass/Agricultural Land and bare land, also effectively classifying small and intricate patterns. Classifying Coastal areas can be difficult in our study area using Landsat-8 due to its resolution. The algorithm well-classified the Coastal areas from all locations. The dense build- up area was well-classified. Taking the Nadi airport region as an example, the comparison between the satellite image and our land cover map of 2013 is shown in the figure. 

\begin{figure*}[!htb]
  \centering
  \medskip
    \begin{subfigure}[t]{.3\linewidth}
    \centering\includegraphics[width=1\linewidth]{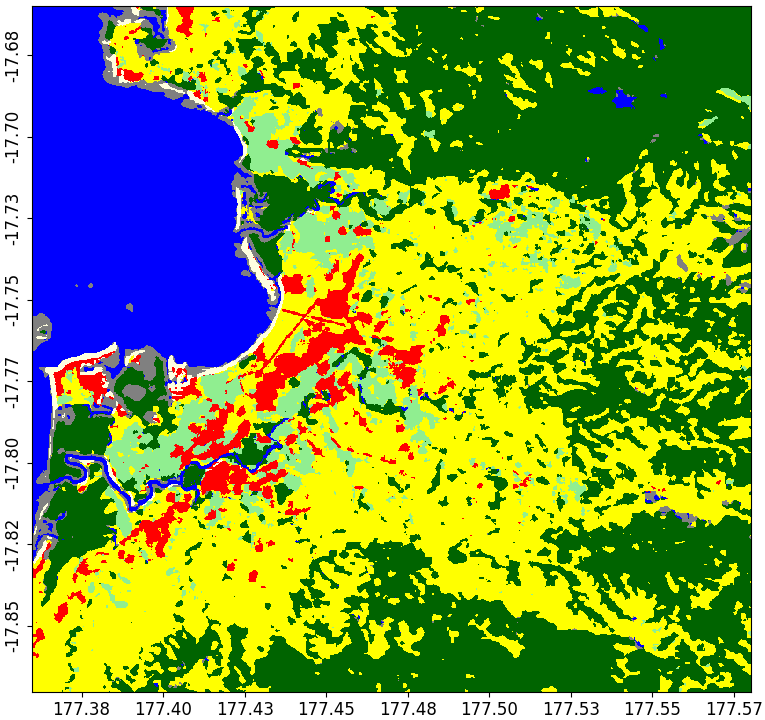}
    \caption{2014}
    \end{subfigure}
    \begin{subfigure}[t]{.3\linewidth}
    \centering\includegraphics[width=1\linewidth]{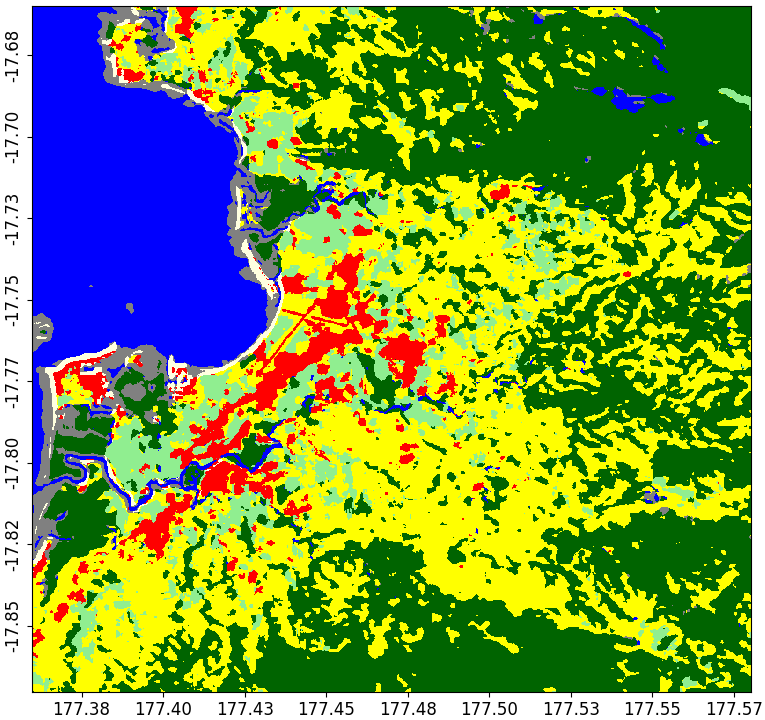}
    \caption{2015}
    \end{subfigure}
    \begin{subfigure}[t]{.3\linewidth}
    \centering\includegraphics[width=1\linewidth]{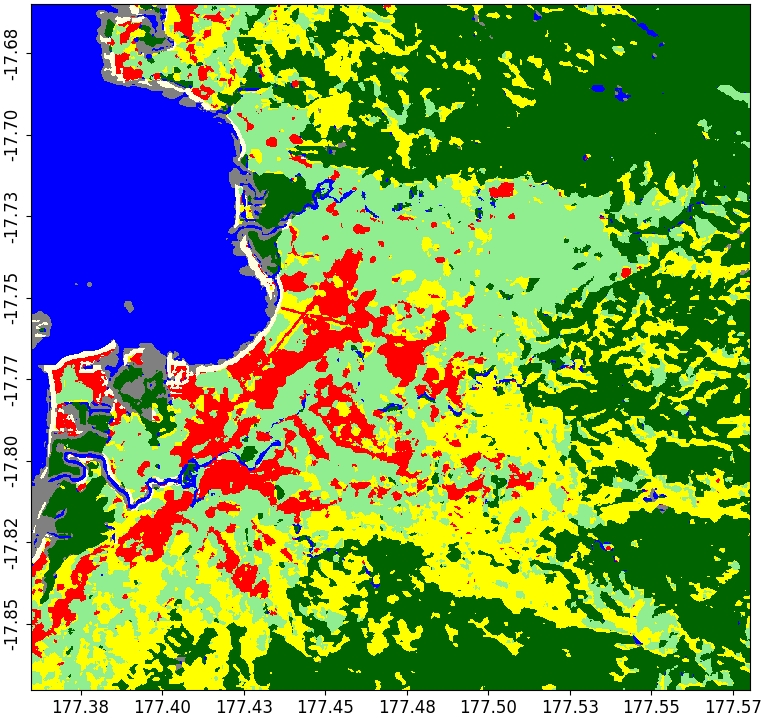}
    \caption{2016}
    \end{subfigure} 
    \begin{subfigure}[t]{.3\linewidth}
    \centering\includegraphics[width=1\linewidth]{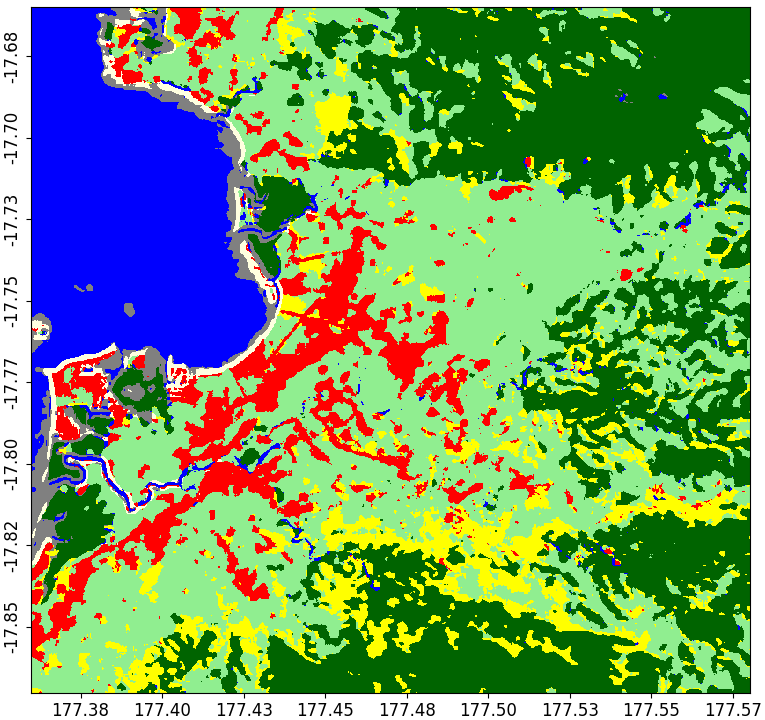}
    \caption{2017}
    \end{subfigure}
    \begin{subfigure}[t]{.3\linewidth}
    \centering\includegraphics[width=1\linewidth]{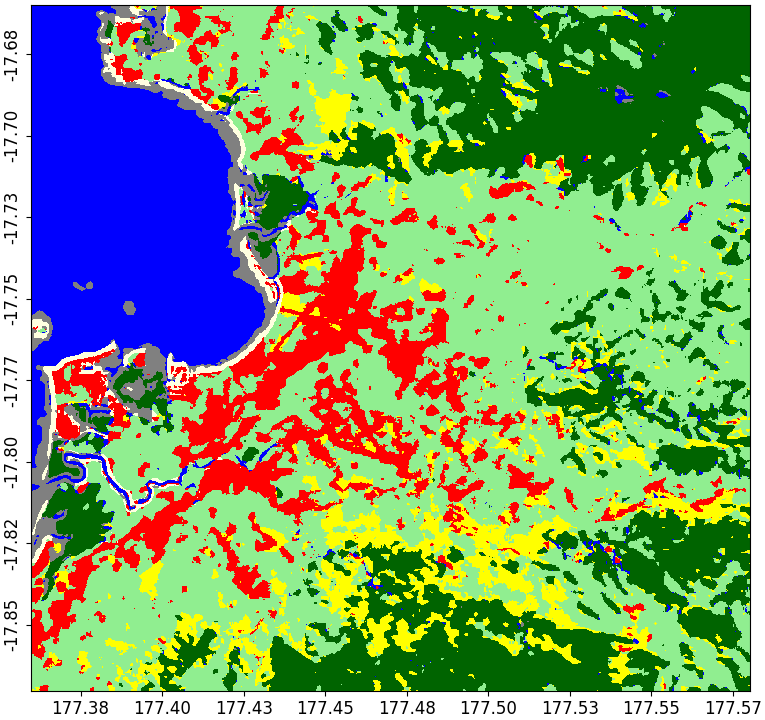}
    \caption{2018}
    \end{subfigure}
    \begin{subfigure}[t]{.3\linewidth}
    \centering\includegraphics[width=1\linewidth]{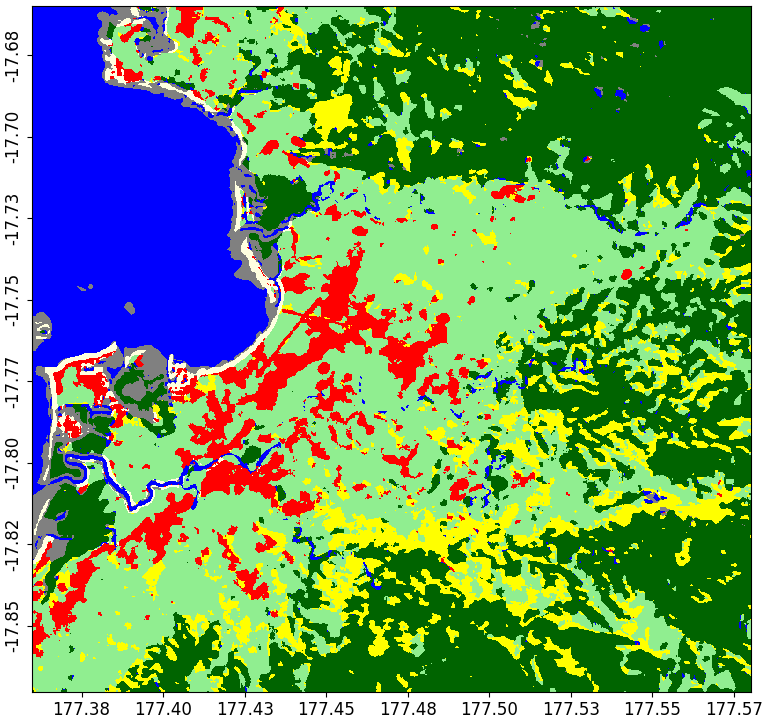}
    \caption{2019}
    \end{subfigure}
    \begin{subfigure}[t]{.3\linewidth}
    \centering\includegraphics[width=1\linewidth]{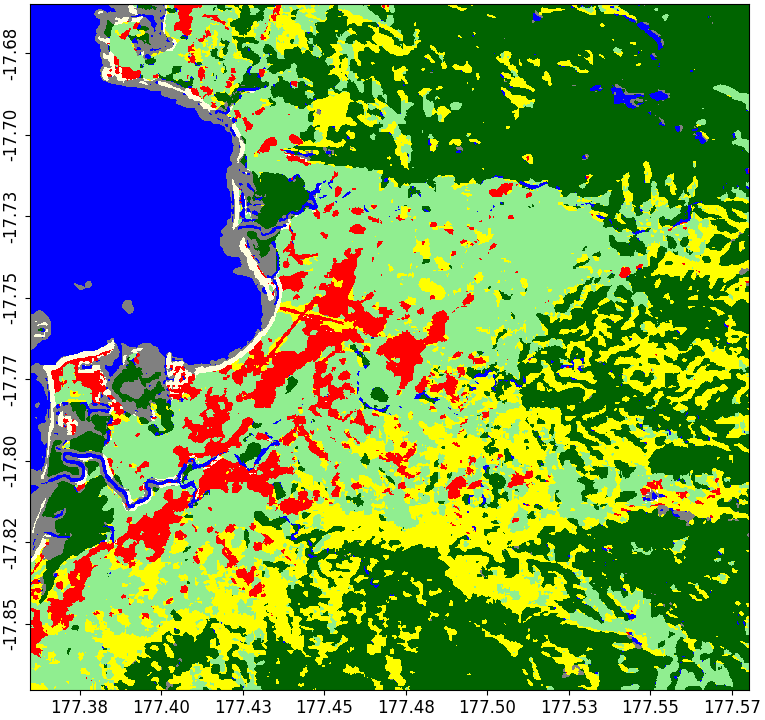}
    \caption{2020}
    \end{subfigure} 
    \begin{subfigure}[t]{.3\linewidth}
    \centering\includegraphics[width=1\linewidth]{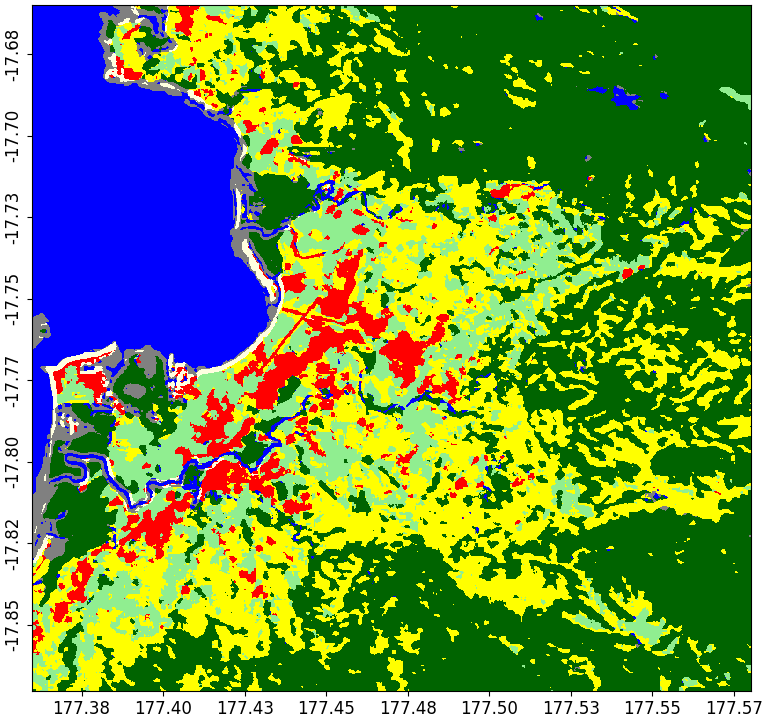}
    \caption{2021}
    \end{subfigure}
    \begin{subfigure}[t]{.3\linewidth}
    \centering\includegraphics[width=1\linewidth]{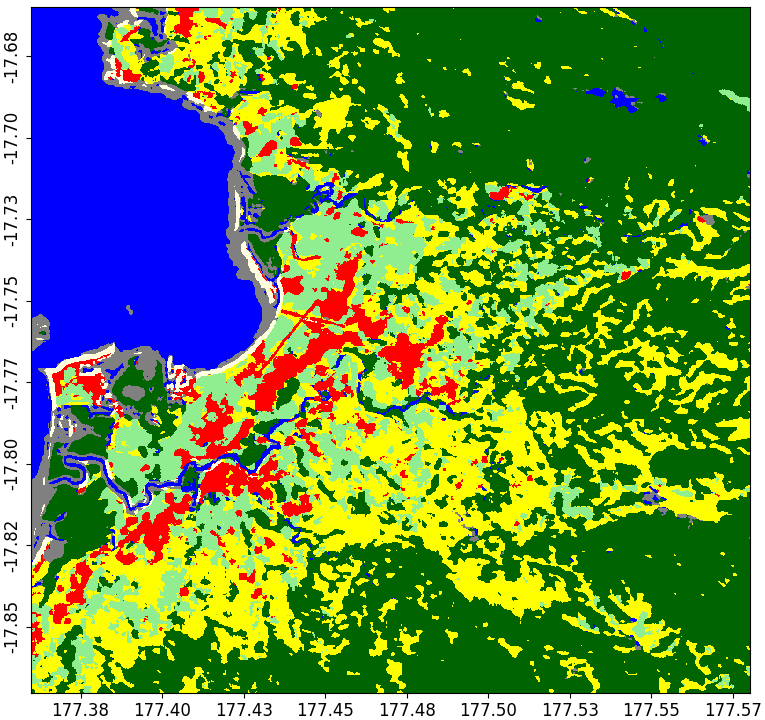}
    \caption{2022}
    \end{subfigure}
    \begin{subfigure}[t]{.3\linewidth}
    \centering\includegraphics[width=1\linewidth]{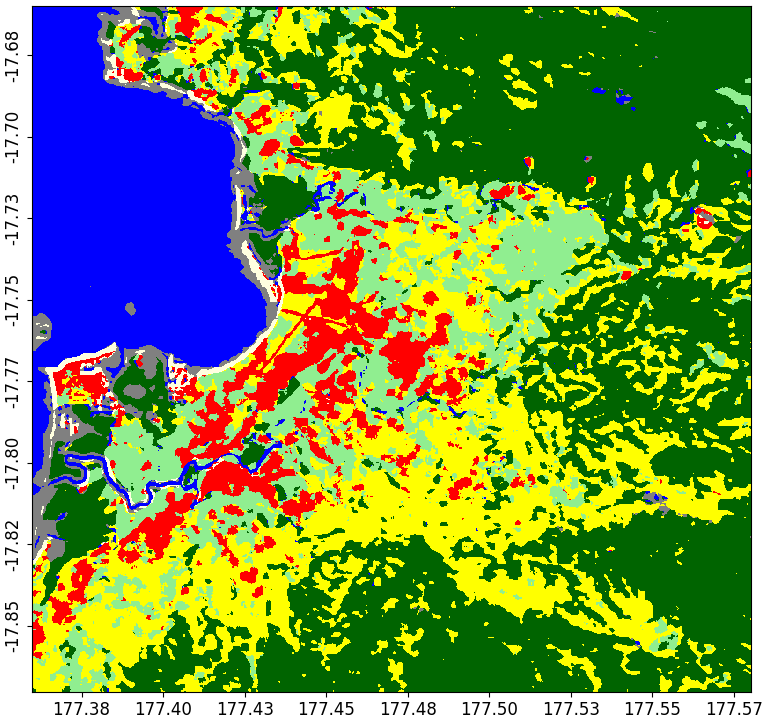}
    \caption{2023}
    \end{subfigure}
    \begin{subfigure}[t]{.6\linewidth}
    \centering\includegraphics[width=1\linewidth]{Class_type.png}
    \end{subfigure} 
\caption{Classified Images from 2014 to 2023}
\end{figure*}

\subsection{Urban Area change detection}

\begin{figure}[!htb]
\includegraphics[width=0.45\textwidth]{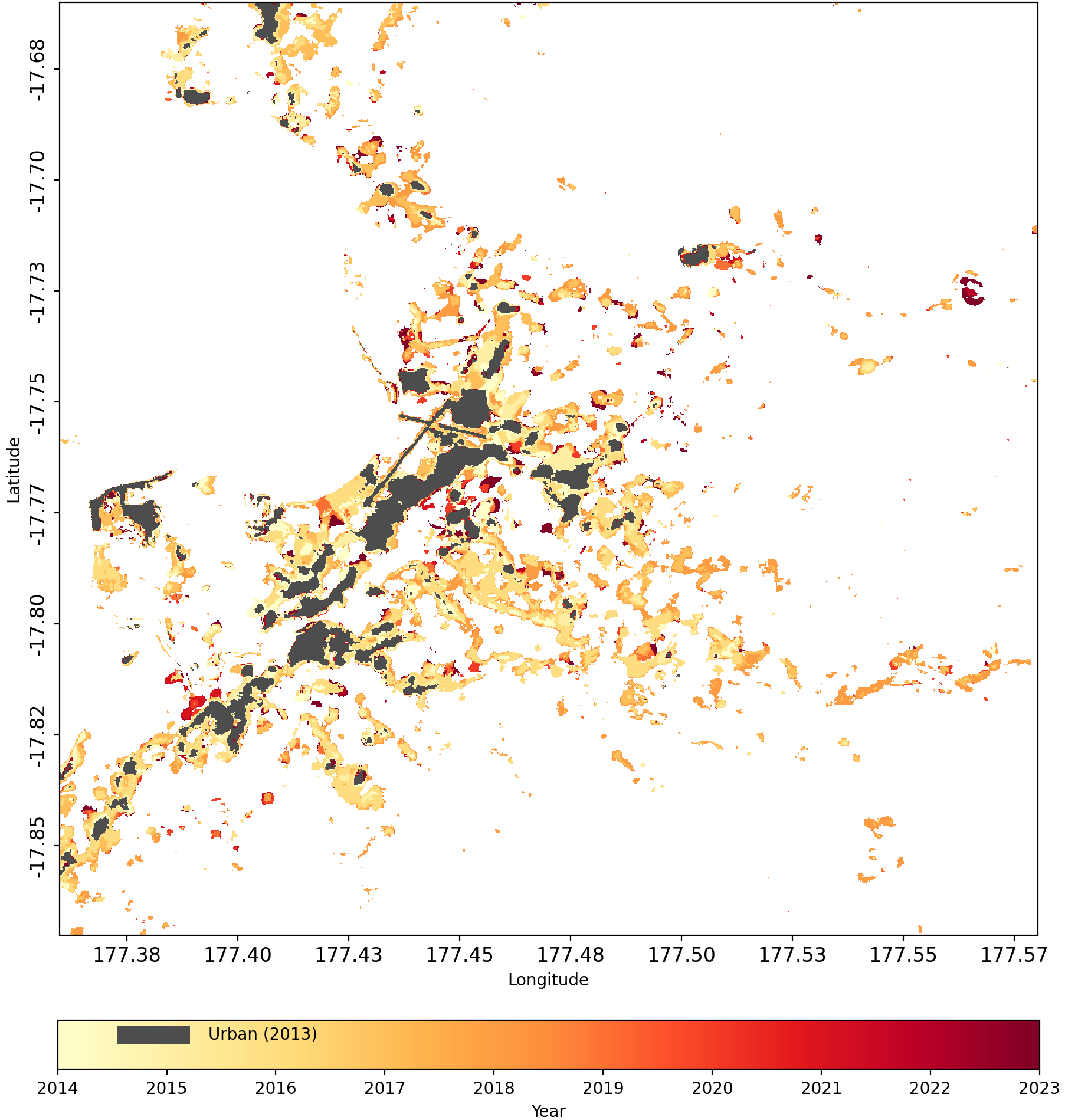}
\caption{Urban Expension Map}
\label{urban_expension_map}
\end{figure}

Based on the land cover maps that were generated by CNN, we created an urban expansion map for visualising the change in the urban area. The urban expansion map Figure~\ref{urban_expension_map} shows how the urban area was expanded from 2013 to 2023. In the image, the non-urban area is represented by the white color.

The urban expansion map reveals that the urban area in our study has expanded outward along the existing periphery and the growth of the urban area in the Nadi downtown, near the coastal line, corresponds to a substantial reduction in grass and agricultural Land coverage. In contrast, a decrease in bare land coverage is observed in the rural area near the mountains. By checking the urban expansion map, it is worth noting that the mangrove areas in Nadi Bay have gradually been replaced by the urban area since 2016.

\section{Discussion}
\label{sec:sample1}
%\subsection{Limitations}

The land cover maps were carefully generated based on each year’s median composite satellite image. The accuracy score for each run was above 98\%, and we showed the detailed change in the urban expansion map and replacement map. Several factors can influence the results.

\noindent\textbf{Sample selection errors: }The land cover maps in this study were generated based on the median composite images of the year, and the purpose was to reduce the impact on classification from haze and clouds of this specific study region. The human error was not avoidable when generating labels on the median composite image of the year by referring to the high-resolution image at a specific time.

 \noindent\textbf{Absence of ground truth data comparison:} A limited amount of official doc- uments or relevant studies are available for comparative analyses of our findings in the study region. Comparing our land cover maps directly with high-resolution ground truth images can not effectively reflect the model’s performance due to the resolution difference. While the assessment score, such as accuracies, precisions etc., are an indispensable criterion when evaluating the model’s classification performance. The random forest case in the previous chapter has demonstrated that high accuracy does not unconditionally ensure the realization of accurate outcomes in the land cover map, especially when the study region is constructed with multiple land cover types.

%\subsection{Future Works}

 %\noindent\textbf{Classification scheme:} 
 
 The classification scheme used in this study described the majority of the land covers in the study area, but there is room for further refinement. The boundary between urban (built-up) and bare soil land cover types is unclear, as shown in the figure 6.1. The sub-image a) and b) are collected from the rural area, and c) is from a construction area collected near Nadi Airport. These regions exhibit distinct signs of human clearing, differentiating them from natural bare land, but their composition remains soil based. Such land cover types are common in many Pacific Island countries and developing countries. This would require a detailed restriction on the boundary or a subdivision of the classes.
 
The Landsat images have been widely used in detecting land cover change. The optical sensor cannot penetrate the clouds, affecting the classification results. When detecting land cover change is required for moist regions like Fiji, data can be collected and combined from different sensors. For example, the data from optical sensors and radar can be used together. This typically requires fusion techniques to integrate the data from different sources.

\section{Conclusion}
\label{sec:sample1}

In this study, we generated land cover maps for Nadi, Fiji and detected the change in urban areas. The current studies lack a unified standard for the number of samples employed in the land cover classification. From comparing 490 samples, 700 samples, 1050 samples, 1400 samples and 1750 samples in the experiments, we found that an appropriate sample size should take up to 0.25\% of the total pixels. We compared three supervised learning algorithms based on the 2013’s satellite image of Nadi, Fiji, including Random Forests, Artificial neural networks, and con- volutional neural networks. The classification accuracies for all three models were high. We also tried K-means on the Google Earth Engine.
While the assessment score, such as accuracies, precisions etc., are the indispensable criterion when evaluating the model’s classification performance. A high accuracy does not unconditionally ensure accurate outcomes in the land cover map, and this typically occurs when the study area encompasses: 1) shades and 2) spectral variation caused by materials and colours within the same land cover class. Random Forest and ANN shared the same input size, yet ANN outperformed. The algorithm’s structure might cause a difference in performance. The pixels fed into the ANN form some connections between layers, whereas random forests take each pixel as a single input, making it unable to learn the spatial characteristics. The CNN algorithm can incorporate spatial context information, leading to outstanding performance in mapping land covers in our study area.
The change detection results indicate that Nadi is experiencing a rapid urbanization process, and the expansion characteristic is that it extends outward along the existing periphery.


\section*{Code and Data}
The code and dataset are publicly available at 
\href{https://github.com/pinglainstitute/landcover-Fiji-}{our GitHub repository}.

 \bibliographystyle{elsarticle-num} 
 \bibliography{cas-refs}





% \end{thebibliography}
\end{document}